\documentclass[letterpaper]{article} 
\usepackage[draft]{aaai2026}  
\usepackage{times}  
\usepackage{helvet}  
\usepackage{courier}  
\usepackage[hyphens]{url}  
\usepackage{graphicx} 
\usepackage{natbib}  
\usepackage{caption} 
\usepackage{algorithm}
\usepackage{algorithmic}

\usepackage{url}            
\usepackage{booktabs}       
\usepackage{amsfonts}       
\usepackage{nicefrac}       
\usepackage{microtype}      
\usepackage{xcolor}         
\usepackage{amsmath}
\usepackage{amssymb}
\usepackage{mathtools}
\usepackage{amsthm}
\usepackage{graphicx}
\usepackage{multirow}
\usepackage{subcaption}
\usepackage{makecell}

\newtheorem{proposition}{Proposition}
\newtheorem{lemma}{Lemma}

\usepackage{newfloat}
\usepackage{listings}
\DeclareCaptionStyle{ruled}{labelfont=normalfont,labelsep=colon,strut=off} 
\floatstyle{ruled}
\newfloat{listing}{tb}{lst}{}
\floatname{listing}{Listing}
\usepackage{hyperref} 

\title{Adapt before Continual Learning}
\author{
    Aojun Lu\textsuperscript{\rm 1}, Tao Feng\textsuperscript{\rm 2}, Hangjie Yuan\textsuperscript{\rm 3}, Chunhui Ding\textsuperscript{\rm 1}, Yanan Sun\textsuperscript{\rm 1}
}
\affiliations{
    \textsuperscript{\rm 1}College of Computer Science, Sichuan University, Chengdu, China \\
    \textsuperscript{\rm 2}Department of Computer Science and Technology, Tsinghua University, Beijing, China\\
    \textsuperscript{\rm 3}College of Computer Science and Technology, Zhejiang University, Hangzhou, China\\
}

\usepackage{bibentry}

\begin{document}

\maketitle

\begin{abstract}
Continual Learning (CL) seeks to enable neural networks to incrementally acquire new knowledge (plasticity) while retaining existing knowledge (stability). Although pre-trained models (PTMs) have provided a strong foundation for CL, existing approaches face a fundamental challenge in balancing these two competing objectives. Current methods typically address stability by freezing the PTM backbone, which severely limits the model's plasticity, particularly when incoming data distribution diverges largely from the pre-training data. Alternatively, sequentially fine-tuning the entire PTM can adapt to new knowledge but often leads to catastrophic forgetting, highlighting the critical stability-plasticity trade-off in PTM-based CL. To address this limitation, we propose \textbf{A}dapting PTMs before the core \textbf{CL} process (ACL), a novel framework that introduces a plug-and-play adaptation phase prior to learning each new task. During this phase, ACL refines the PTM backbone by aligning embeddings with their original class prototypes while distancing them from irrelevant classes. This mechanism theoretically and empirically demonstrates desirable balance between stability and plasticity, significantly improving CL performance across benchmarks and integrated methods. Code is available at \url{https://github.com/byyx666/ACL_code}.
\end{abstract}


\section{Introduction}
\label{introduction}
In open-world scenarios, data typically arrives in a streaming fashion, necessitating a machine learning paradigm capable of incrementally acquiring new knowledge while retaining previous information, known as Continual Learning (CL)~\cite{wang2023comprehensive, zhou2024continual}. Effective CL critically hinges on a neural network's ability to balance \textit{plasticity}, which enables the learning of new concepts, and \textit{stability}, which ensures the retention of previously acquired knowledge. 
However, an overemphasis on stability can hinder the model's adaptability to new information, whereas excessive plasticity may lead to \textit{catastrophic forgetting} of prior knowledge~\cite{mccloskey1989catastrophic,goodfellow2013empirical}. This fundamental conflict is known as the \textit{stability-plasticity dilemma}~\cite{grossberg2013adaptive}, which remains a central challenge in CL research.

\begin{figure*}[t]
    \centering
    \subcaptionbox{Plasticity}{\includegraphics[width = 0.32\textwidth]{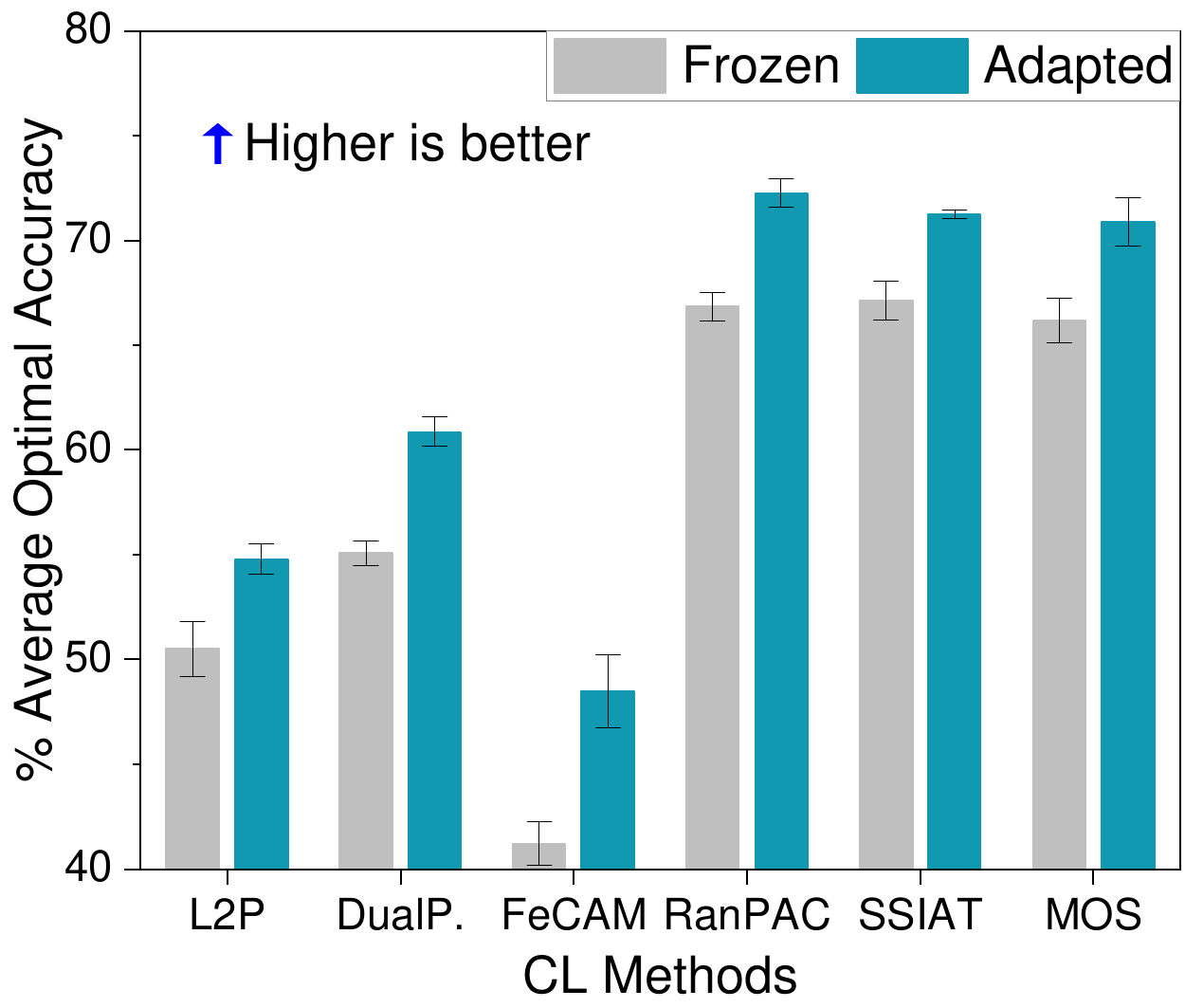}}
    \hfill
    \subcaptionbox{Stability}{\includegraphics[width = 0.32\textwidth]{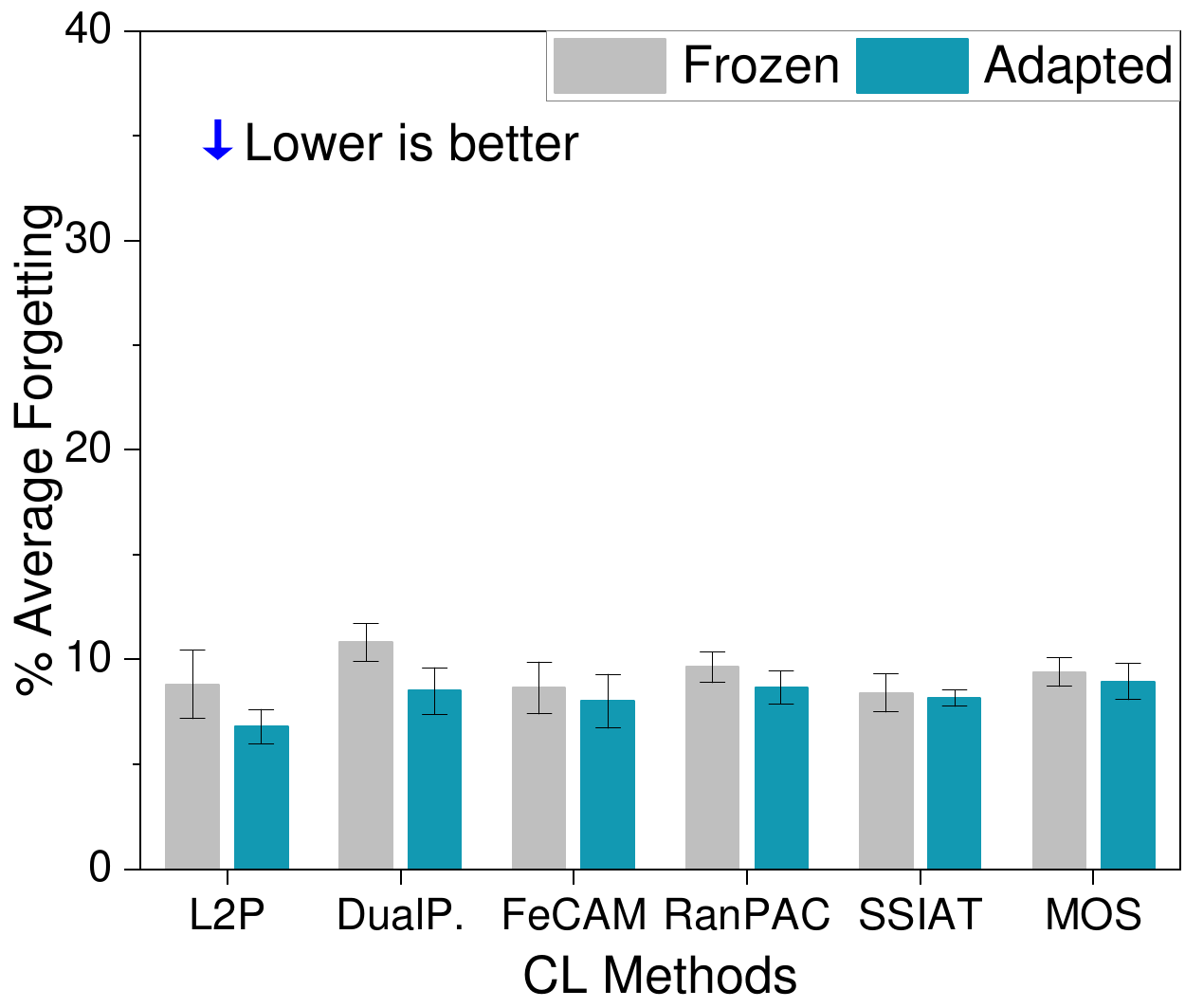}}
    \hfill
    \subcaptionbox{Overall CL performance}{\includegraphics[width = 0.32\textwidth]{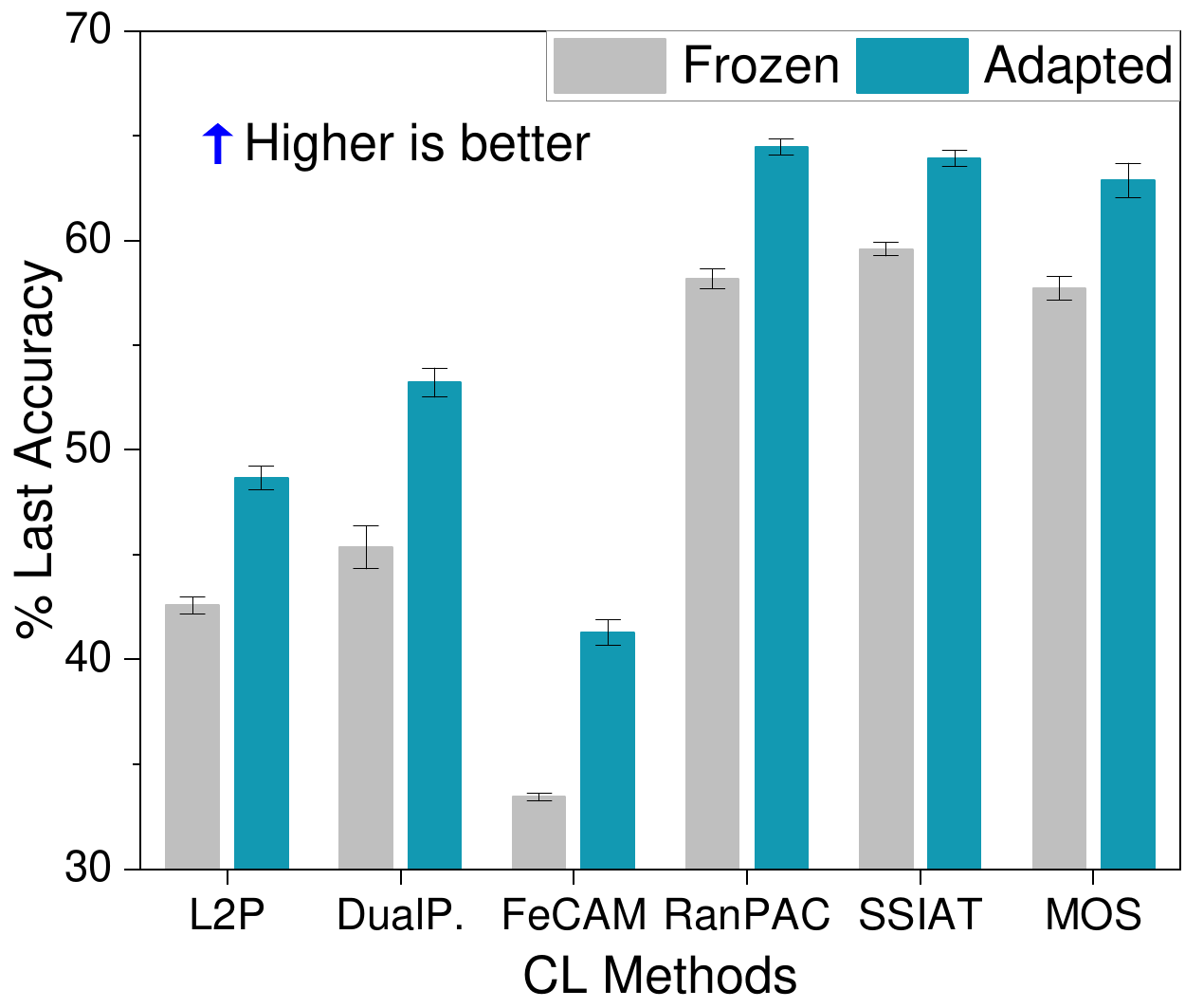}}
\caption{Performance comparison on ImageNet-A-Inc20 between the frozen PTM and the PTM adapted using our ACL. Plasticity: the average of the optimal accuracy of each task during CL; Stability: the average forgetting across previous tasks after learning the final task; Overall CL performance: the average accuracy across all tasks after learning the final task.
}
\label{fig:fig1}
\end{figure*}

The advent of powerful Pre-Trained Models (PTMs) has significantly reshaped the machine learning domain, spurring considerable interest in their application to CL~\cite{zhou2024continual}. PTMs, trained on large-scale datasets, exhibit strong generalization capabilities, providing a strong foundation for downstream CL tasks. A dominant strategy in PTM-based CL, therefore, involves freezing the PTM backbone to preserve this foundational knowledge while training lightweight modules (\textit{e.g.,} prompts or adapters) tailored to new tasks~\cite{wang2022learning, wang2022dualprompt, tan2024semantically, sun2024mos}. While such parameter-efficient approaches excel at preserving PTMs' generalizable knowledge for stability, they may trade off plasticity, as the model cannot sufficiently adapt its learned representations to new tasks.

A critical challenge emerges from this trade-off: the pre-trained feature space, though broadly general, is not always optimal for the discriminative requirements of every downstream task~\cite{zhou2022domain,hendrycks2021natural}. When the incoming data distribution diverges significantly from the pre-training data, the highly stable but rigid feature representations can become a bottleneck, limiting the model's ability to learn new concepts effectively~\cite{zhou2024revisiting}. This limitation may result in diminished plasticity and suboptimal CL performance. On the other hand, fully fine-tuning the PTM for each task may enhance plasticity but risks degrading its generalizable knowledge, leading to catastrophic forgetting~\cite{kumarfine}. This underscores a more nuanced dilemma in PTM-based CL: \textit{how to realign the PTM's feature space for new tasks without destabilizing the foundational knowledge crucial for stability.}

To address this pivotal challenge, we introduce \textbf{A}dapting PTMs before the core \textbf{CL} process (ACL), a novel framework that performs a plug-and-play adaptation of the PTM's feature space before learning each new task. By adapting the PTM's feature space to align better with incremental data, ACL provides a stronger and more task-relevant foundation for existing PTM-based CL methods to build upon. Specifically, during the adaptation phase, the PTM backbone is fine-tuned by encouraging its output embeddings to move closer to their respective original class prototypes while simultaneously distancing them from other class prototypes. This straightforward yet effective adaptation mechanism ensures that the PTM's feature space is realigned for new tasks without destabilizing the foundational knowledge critical for stability. As illustrated in Fig.~\ref{fig:fig1}, while existing CL methods using frozen PTMs exhibit strong stability (with approximately 10\% forgetting), their overall CL performance remains limited by weak plasticity. Integrating ACL not only significantly enhances plasticity but also preserves stability, yielding superior CL performance. 

The contributions of this study are outlined as follows: (i) We identify that prevailing PTM-based CL methods achieve suboptimal CL performance due to inherent limitations in plasticity, underscoring the necessity for an effective adaptation mechanism. (ii) We demonstrate theoretically the core objectives of adaptation, \textit{i.e.}, enhancing plasticity while preserving stability, can be effectively achieved by encouraging embeddings to converge toward their original class prototypes and diverge from others. (iii) We propose ACL, a novel, plug-and-play CL framework designed to achieve a improved stability-plasticity trade-off. Extensive experiments across diverse benchmarks and established CL methods validate the effectiveness and broad applicability of ACL.

\section{Related Works}
\label{related_works}

\subsection{Continual Learning (CL)}

CL aims to enable neural networks to sequentially acquire knowledge from a series of tasks without forgetting previously learned concepts~\cite{masana2022class,van2022three}. Traditional CL methods can be broadly categorized into three types. \textit{Replay-based} methods retain a subset of previous data information in a memory buffer, which is subsequently utilized to recover old data distributions~\cite{aljundi2019gradient, liu2020mnemonics, iscen2020memory, zhao2021memory}. \textit{Regularization-based} methods incorporate penalty terms that constrain model updates during the learning of new tasks~\cite{kirkpatrick2017overcoming, zenke2017continual, li2017learning, feng2022overcoming}. \textit{Architecture-based} methods allocate task-specific parameter spaces within the network for each new task, thereby mitigating forgetting~\cite{kang2022forget, konishi2023parameter, yan2021dynamically, zhou2023model}.

\textbf{CL with PTMs.} With the growing prevalence of PTMs~\cite{dosovitskiy2020image, radford2021learning}, PTM-based CL has recently garnered significant attention. Given that PTMs have been equipped with generalizable knowledge, these methods often freeze the pre-trained backbones and utilize additional trainable modules to learn task-specific knowledge~\cite{zhou2024continual}. Early research primarily focuses on applying visual prompt tuning~\cite{jia2022visual} to CL, enabling models to learn new tasks without modifying the pre-trained weights~\cite{smith2023coda, jung2023generating}, \textit{e.g.,} L2P~\cite{wang2022learning} and DualPrompt~\cite{wang2022dualprompt}. Recently, some studies have demonstrated that adapter-based tuning outperforms prompt-based methods in PTM-based CL~\cite{tan2024semantically, gao2024beyond}, \textit{e.g.,} SSIAT~\cite{tan2024semantically} and MOS~\cite{sun2024mos}. In addition to developing additional trainable modules, several studies have focused on optimizing the classification head to enhance CL performance, \textit{e.g.,} FeCAM~\cite{goswami2024fecam} and RanPAC~\cite{mcdonnell2024ranpac}.

\subsection{Prototypical Networks}

Prototypical networks~\cite{snell2017prototypical} involve learning an embedding space where samples are classified by minimizing their distance to the mean embedding (prototype) of their respective class~\cite{li2020prototypical,zhang2022hierarchical}. In our research, we demonstrate, both theoretically and empirically, that the integration of this core principle into the adaptation phase of PTMs achieves a desirable balance between plasticity and stability.

\subsection{Contrastive Learning}

Contrastive learning~\cite{oord2018representation} has emerged as a powerful framework in self-supervised learning and supervised learning~\cite{khosla2020supervised}, which brings similar examples closer together in the feature space while pushing dissimilar examples apart. In the context of CL, several studies have leveraged contrastive learning to enhance stability~\cite{nagata2023margin, wen2024provable}, \textit{e.g.,} Co$^2$L~\cite{cha2021co2l} and PCL~\cite{lin2023pcr}. These approaches typically contrast embeddings from current task data against replayed samples from previous tasks (exemplar replay) to preserve learned representations. Unlike these methods that primarily utilize contrastive learning with replayed data to bolster stability, our work focuses on applying contrastive principles exclusively to the current task's data. Furthermore, our primary objective through this application is to enhance plasticity, with stability being an emergent benefit of our formulation.

\section{ACL: Adapt before Continual Learning}
\label{methods}

To enhance plasticity in CL with PTMs, we propose a novel CL framework, ACL, which introduces a novel adaptation phase before learning each incremental task. This phase aims to adapt the PTM's weights to the characteristics of the new data, thereby improving feature discriminability for the current task, while preserving the existing knowledge. In the subsequent sections, we first outline the overall procedure of ACL. We then detail the specific adaptation loss function employed and provide a theoretical analysis demonstrating how it addresses the stability-plasticity trade-off.

\subsection{Preliminaries} 

For clarity, we decompose the CL model into two primary components: $f(x)=\mathcal{C}(\phi(x))$. Here, $\phi(\cdot): \mathbb{R}^{D} \rightarrow \mathbb{R}^{d}$ represents the PTM backbone, which maps input samples $x$ into feature embeddings. The classification head, $\mathcal{C}(\cdot): \mathbb{R}^{d} \rightarrow \mathbb{R}^{|\mathcal{Y}_k|}$, takes these embeddings and produces classification outputs. Existing PTM-based CL approaches often freeze the backbone $\phi(\cdot)$ throughout the learning process to preserve this general knowledge. To acquire knowledge from new tasks, these methods typically introduce additional trainable modules, denoted by $\Theta$ (\textit{e.g.,} prompts and adapters). The model then becomes $f(x)=\mathcal{C}(\phi(x),\Theta)$, and only $\Theta$ and the classification head $\mathcal{C}$ are updated during CL.

\subsection{Overall Procedure of ACL.}

\begin{figure}[ht]
    \centering
    {\includegraphics[width = 0.96\linewidth]{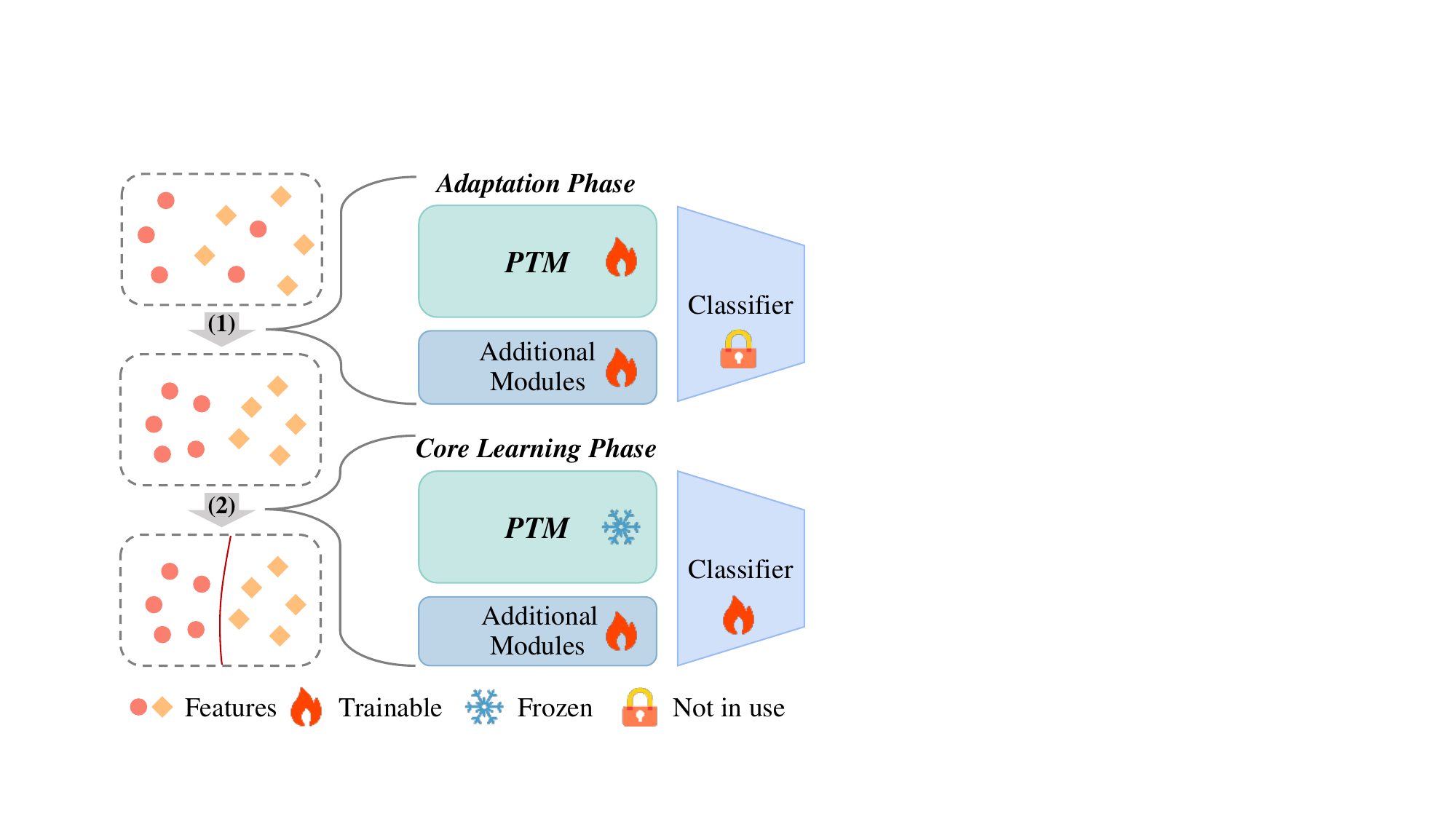}}
    \caption{Illustration of ACL. ACL comprises two phases per task: (1) Adapting the PTM weights to enhance feature discriminability for the current task, and (2) Learning classification using the frozen adapted PTM and trainable modules.
}
\label{fig:framework}
\end{figure}
 
The ACL framework operates in two distinct phases for each new task, as illustrated in Fig.~\ref{fig:framework}. 

\textbf{Phase 1: Adaptation.} At the beginning of learning the $k$-th task, given the data $\mathcal{D}_{k}$, we adapt the parameters of the PTM backbone $\phi_{k-1}$ and the lightweight modules $\Theta_{k-1}$. This adaptation aims to make the features generated by the model more discriminative for the current task. Formally, this process is defined as:
\begin{equation}
\label{eq:adapt}
\phi_{k-1}^{*}, \Theta_{k-1}^{*} = \mathcal{A}(\phi_{k-1}, \Theta_{k-1}, \mathcal{D}_{k}),
\end{equation}
where $\mathcal{A}$ denotes the adaptation algorithm. The output consists of the adapted backbone $\phi_{k-1}^{*}$ and adapted lightweight modules $\Theta_{k-1}^{*}$. Critically, the design of $\mathcal{A}$ ensures that the separability of features for $\mathcal{D}_{k}$ is enhanced while attempting to maintain the existing knowledge encoded in $\phi_{k-1}$ and $\Theta_{k-1}$. The specific design of $\mathcal{A}$ is detailed in the next section.

\textbf{Phase 2: Core Learning.} Following adaptation, the adapted backbone $\phi^{*}_{k-1}$ is frozen. Subsequently, the classification head $\mathcal{C}_{k-1}$ and the adapted lightweight modules $\Theta^{*}_{k-1}$ are fine-tuned to learn the classification task for $\mathcal{D}_{k}$. This step leverages the adapted PTM's knowledge for the new task. The process is formalized as:
\begin{equation}
\label{eq:learn}
    \begin{aligned}
    \phi_{k} &= \phi^{*}_{k-1}, \\
    \mathcal{C}_{k}, \Theta_{k} &= \mathcal{F}(\mathcal{C}_{k-1}, \Theta_{k-1}^{*}, \phi_{k}, \mathcal{D}_{k}),
    \end{aligned}
\end{equation}
where $\mathcal{F}$ represents the CL method integrated into the ACL framework, \textit{e.g.}, L2P~\cite{wang2022learning}.

Algorithm~\ref{alg:ACL} presents the pseudo-code for ACL. It iteratively applies the adaptation (Eq.~\ref{eq:adapt}) and core learning (Eq.~\ref{eq:learn}) phases for each incremental task from $k=1$ to $K$.

\begin{algorithm}[ht]  
    \caption{Procedure of ACL} 
    \label{alg:ACL}
\begin{algorithmic}[1]
    \STATE {\bfseries Input:} PTM backbone $\phi_0$, Additional lightweight module $\Theta_0$, Classification Head $\mathcal{C}_0$, Incremental datasets $\{\mathcal{D}_1, \mathcal{D}_2, \ldots, \mathcal{D}_K\}$;
    \STATE {\bfseries Output:} Updated PTM;
    \FOR{task $k = 1, 2, \dots, K$}    
        \STATE Get the training set of the incremental dataset $\mathcal{D}_{k}$;
        
        \STATE Optimize $\phi_{k-1}$ and $\Theta_{k-1}$ to obtain $\phi^{*}_{k-1}$ and $\Theta^{*}_{k-1}$ via Eq.~\ref{eq:adapt};
        
        \STATE Save $\phi^{*}_{k-1}$, \textit{i.e.}, $\phi_{k}=\phi^{*}_{k-1}$;
        
        \STATE Optimize $\Theta^{*}_{k-1}$ and $\mathcal{C}_{k-1}$ to obtain $\Theta_{k}$ and $\mathcal{C}_{k}$ via Eq.~\ref{eq:learn} with frozen $\phi_{k}$;
    \ENDFOR
\end{algorithmic}
\end{algorithm} 

\subsection{Adaptation Algorithm within ACL}

This section introduces the adaptation loss function used within the $\mathcal{A}$ phase of ACL, termed the ACL loss. This loss is specifically designed to navigate the stability-plasticity trade-off inherent in CL. We provide a theoretical justification showing that the ACL loss simultaneously (1) promotes \textit{plasticity} by minimizing an upper bound on the current task's classification error and (2) maintains \textit{stability} by implicitly regularizing feature deviation.

For analytical simplicity, we consider a cosine classifier, a common choice in PTM-based CL~\cite{zhou2024revisiting, sun2024mos}. This classifier assigns a sample $x$ to the class $c$ whose prototype $p_c$ has the highest cosine similarity with the feature embedding. We note that, although the theoretical analysis is based on the cosine classifier, the empirical results demonstrate the applicability of ACL to CL methods using other classifiers, \textit{e.g.}, the linear classifier.

\subsubsection{ACL Loss}
Let $\phi$ be the PTM backbone prior to adaptation and $\phi^{*}$ the backbone after adaptation. For the current task $\mathcal{D}_k$, the adaptation process computes class prototypes $p_c=\mathbb{E}_{(x,y)\in\mathcal{D}_k,\,y=c}[\phi(x)]$ for every class $c$. Note that all feature embeddings $\phi(x),\phi^{*}(x)$ and prototypes $p_c$ are $\ell_2$-normalized to the unit hypersphere $\mathbb{S}^{d-1}$. The ACL loss encourages high cosine similarity between an adapted embedding and its corresponding class prototype, while penalizing similarities to incorrect prototypes:

\begin{equation}
\mathcal{L}_{\text{ACL}}(x_i,y_i)=-\log\frac{\exp \bigl(\cos(\phi^{*}(x_i),p_{y_i})/\tau\bigr)}{\sum_j \exp \bigl(\cos(\phi^{*}(x_i),p_j)/\tau\bigr)},
\end{equation}
where $\tau$ is a temperature parameter.

\subsubsection{Plasticity Analysis}
We first establish that minimizing the ACL loss promotes plasticity by directly reducing an upper bound on the classification error for the current task.

\begin{proposition}
For the adapted model $\phi^{*}(\cdot)$, the probability of misclassifying a sample $x_i$ is upper-bounded by the expected ACL loss:
\begin{equation}
P(\text{misclassify})\;\le\;\frac{\mathbb{E}[\mathcal{L}_{\text{ACL}}]}{\log 2}.
\end{equation}
\end{proposition}

\begin{proof}
By rewriting the ACL loss in terms of log-sum-exp, we obtain:
\begin{equation}
    \begin{aligned}
    \mathcal{L}_{\text{ACL}}(x_i, y_i) 
    &= -\log \frac{\exp(S_{y_i})}{\exp(S_{y_i}) + \sum_{j \neq y_i} \exp(S_j)} \\
    &= \log \frac{\exp(S_{y_i}) + \sum_{j \neq y_i} \exp(S_j)}{\exp(S_{y_i})} \\
    &= \log \bigl(1 + \sum_{j \neq y_i} \exp(S_j - S_{y_i})\bigr).
    \end{aligned}
\end{equation}
\noindent where $S_{j} = \cos(\phi^{*}(x_i), p_j)/\tau$; $S_{y_i} = \cos(\phi^{*}(x_i), p_{y_i})/\tau$. 

By definition, for the adapted model $\phi^{*}$ using a cosine classifier, a sample $(x_i, y_i)$ is misclassified if and only if there exists an incorrect class $k \neq y_i$ such that $\cos(\phi(x_i), p_k) \ge \cos(\phi(x_i), p_{y_i})$. Therefore, in the event of misclassification, there exists at least one $k$ such that $S_{k} - S_{y_i} \ge 0$, making $\exp(S_k - S_{y_i}) \ge 1$. Thus, the sum $\sum_{j \neq y_i} \exp(S_j - S_{y_i}) \ge 1$, and $\mathcal{L}_{\text{ACL}}(x_i, y_i) \ge \log(1 + 1) = \log 2$. That is, $\text{misclassify} \Longrightarrow \mathcal{L}_{\text{ACL}}(x_i,y_i) \ge \log(1 + 1) = \log 2$.

Therefore, $P(\text{misclassify}) \le P(\mathcal{L}_{\text{ACL}} \ge \log 2)$. Applying Markov's inequality to $\mathcal{L}_{\text{ACL}}$ yields the final bound:
\begin{equation}
P(\text{misclassify})\;\le\;P \bigl(\mathcal{L}_{\text{ACL}}\ge\log 2\bigr)\;\le\;\frac{\mathbb{E}[\mathcal{L}_{\text{ACL}}]}{\log 2}.
\end{equation}
\end{proof}

\textbf{Conclusion on plasticity.} Minimizing the expected ACL loss directly minimizes a probabilistic upper bound on the current-task classification error, thereby enhancing plasticity.

\subsubsection{Stability Analysis}
We now demonstrate that the ACL loss maintains stability through an implicit regularization mechanism that constrains feature deviation.

\begin{lemma}
\label{le:1}
For any two $\ell_2$-normalized vectors $a, b \in \mathbb{S}^{d-1}$, the squared Euclidean distance is directly related to their cosine similarity:
\begin{equation}
\|a - b\|_2^2 = 2\bigl(1 - \cos(a, b)\bigr).
\end{equation}
\end{lemma}

\begin{lemma}
\label{le:2}
For a give class $y_{i}$, the prototype is the unique point that minimizes the expected squared Euclidean distance to all features.
\begin{equation}
p_{y_{i}}=\underset{z}{\arg \min } \mathbb{E}_{(x, y) \in \mathcal{D}_{k, y=y_{i}}}\left[\|\phi(x)-z\|_{2}^{2}\right].
\end{equation}
\end{lemma}

\begin{proposition}  
The ACL loss implicitly regularizes the feature deviation. Specifically, it enforces a tight upper bound on the expected feature deviation $\mathbb{E}[\|\phi^{*}(x) - \phi(x)\|_2^2]$.
\end{proposition}

\begin{proof}  
Consider the expected squared feature change over the current task distribution $\mathcal{D}_k$:
$$
\mathbb{E}_{(x,y)\sim\mathcal{D}_k}[\|\phi^{*}(x) - \phi(x)\|_2^2].
$$

For any sample $(x_i, y_i)$ from $\mathcal{D}_k$, we can bound the squared feature change $\|\phi^{*}(x_i) - \phi(x_i)\|_2^2$ by using the class prototype $p_{y_i}$ as an intermediate point and applying a fundamental inequality for squared norms:
\begin{equation}
    \|\phi^{*}(x_i) - \phi(x_i)\|_2^2 \le 2 ({\|\phi^{*}(x_i) - p_{y_i}\|_2^2} + {\|\phi(x_i) - p_{y_i}\|_2^2}).
\end{equation}

Taking expectations over $x \sim \mathcal{D}_k$ on both sides preserves the inequality due to the linearity of expectation:
\begin{equation}
    \begin{aligned}
    &\mathbb{E}_{(x,y)\sim\mathcal{D}_k}[\|\phi^{*}(x) - \phi(x)\|_2^2] \le \\ &2\left(\mathbb{E}_{(x,y)\sim\mathcal{D}_k}[\|\phi^{*}(x) - p_y\|_2^2] + \mathbb{E}_{(x,y)\sim\mathcal{D}_k}[\|\phi(x) - p_y\|_2^2]\right).
    \end{aligned}
\end{equation}

Let us now analyze each expected term:

\begin{itemize}
    \item $\mathbb{E}_{(x,y)\sim\mathcal{D}_k}[\|\phi^{*}(x) - p_y\|_2^2]$. By the Lemma~\ref{le:1}, this equals $2\,\mathbb{E}_{(x,y)\sim\mathcal{D}_k}[1 - \cos(\phi^{*}(x), p_y)]$. ACL loss directly minimizes this term by maximizing the expected cosine similarity $\mathbb{E}_{(x,y)\sim\mathcal{D}_k}[\cos(\phi^{*}(x), p_y)]$.

    \item $\mathbb{E}_{(x,y)\sim\mathcal{D}_k}[\|\phi(x) - p_y\|_2^2]$. Since $\phi$ is fixed during adaptation and the prototypes $p_y$ are computed from $\phi$, this term is constant with respect to the adaptation process. Moreover, by Lemma~\ref{le:2}, the use of class prototypes ensures that this term is minimized over all possible anchors, thereby tightening the overall bound.
\end{itemize}

As a result, the ACL loss serves as an implicit regularization mechanism that constrains feature deviation.
\end{proof}

\textbf{Conclusion on stability.} Minimizing the ACL loss implicitly constrains the deviation between original and adapted features, thereby maintaining stability. Intuitively, this is achieved by anchoring the adapted features to the most representative points in the original feature space, \textit{i.e.}, prototypes.

\begin{table*}[ht]
    \caption{Performance (\%) of six state-of-the-art CL methods with/without ACL. `Improvement' represents the boost of ACL.}
    \label{tab:result_integration}
    \centering
    \begin{tabular}{@{\hspace{2mm}}lcccccccc@{\hspace{2mm}}}
        \toprule
        \multirow{2.4}*{Method} &\multicolumn{2}{c}{ImageNet-R-Inc20} &\multicolumn{2}{c}{ImageNet-R-Inc10} &\multicolumn{2}{c}{ImageNet-A-Inc20} &\multicolumn{2}{c}{ImageNet-A-Inc10}\\
        \cmidrule(lr){2-9}
        &LA &AIA &LA &AIA &LA &AIA &LA &AIA\\

        \midrule
        L2P &71.91\scriptsize{$\pm$0.27} &76.76\scriptsize{$\pm$0.45} &69.24\scriptsize{$\pm$0.78} &74.61\scriptsize{$\pm$0.61} &42.58\scriptsize{$\pm$0.39} &50.42\scriptsize{$\pm$1.12} &34.93\scriptsize{$\pm$0.96} &44.24\scriptsize{$\pm$1.25}\\
        w/ Ours &75.47\scriptsize{$\pm$0.53} &80.09\scriptsize{$\pm$0.40} &73.07\scriptsize{$\pm$0.57} &78.49\scriptsize{$\pm$0.56} &48.65\scriptsize{$\pm$0.55} &55.01\scriptsize{$\pm$1.24} &41.92\scriptsize{$\pm$1.61} &49.34\scriptsize{$\pm$2.06}\\
        \cmidrule(lr){1-9}
        Improvement &\textbf{+3.56} &\textbf{+3.33} &\textbf{+3.83} &\textbf{+3.88} &\textbf{+6.07} &\textbf{+4.59} &\textbf{+6.99} &\textbf{+5.10}\\
        
        \midrule
        DualPrompt &69.43\scriptsize{$\pm$0.51} &74.85\scriptsize{$\pm$0.18} &65.71\scriptsize{$\pm$0.24} &71.89\scriptsize{$\pm$0.34} &45.35\scriptsize{$\pm$1.04} &54.72\scriptsize{$\pm$1.64} &39.04\scriptsize{$\pm$1.83} &49.46\scriptsize{$\pm$2.26}\\
        w/ Ours &74.97\scriptsize{$\pm$0.25} &79.88\scriptsize{$\pm$0.44} &72.07\scriptsize{$\pm$0.32} &77.52\scriptsize{$\pm$0.40} &53.22\scriptsize{$\pm$0.70} &60.02\scriptsize{$\pm$1.91} &46.65\scriptsize{$\pm$1.19} &54.97\scriptsize{$\pm$2.01}\\
        \cmidrule(lr){1-9}
        Improvement &\textbf{+5.54} &\textbf{+5.03} &\textbf{+6.36} &\textbf{+5.63} &\textbf{+7.87} &\textbf{+5.30} &\textbf{+7.61} &\textbf{+5.51}\\
        
        \midrule
        FeCAM &60.39\scriptsize{$\pm$1.30} &66.15\scriptsize{$\pm$1.24} &55.60\scriptsize{$\pm$0.26} &61.98\scriptsize{$\pm$0.42} &33.43\scriptsize{$\pm$0.18} &41.89\scriptsize{$\pm$0.95} &33.79\scriptsize{$\pm$0.10} &42.96\scriptsize{$\pm$0.65}\\
        w/ Ours &65.82\scriptsize{$\pm$0.80} &70.33\scriptsize{$\pm$0.86} &63.05\scriptsize{$\pm$1.48} &67.96\scriptsize{$\pm$1.25} &41.28\scriptsize{$\pm$0.61} &46.67\scriptsize{$\pm$1.89} &38.62\scriptsize{$\pm$0.44} &45.56\scriptsize{$\pm$0.98}\\
        \cmidrule(lr){1-9}
        Improvement &\textbf{+5.43} &\textbf{+4.18} &\textbf{+7.45} &\textbf{+5.98} &\textbf{+7.85} &\textbf{+4.78} &\textbf{+4.83} &\textbf{+2.60}\\
        
        \midrule
        RanPAC &76.07\scriptsize{$\pm$0.85} &81.18\scriptsize{$\pm$0.94} &72.84\scriptsize{$\pm$0.23} &78.47\scriptsize{$\pm$0.55} &58.16\scriptsize{$\pm$0.46} &66.73\scriptsize{$\pm$1.47} &57.33\scriptsize{$\pm$1.26} &65.79\scriptsize{$\pm$1.55}\\
        w/ Ours &79.14\scriptsize{$\pm$0.21} &83.29\scriptsize{$\pm$0.50} &78.20\scriptsize{$\pm$0.25} &82.37\scriptsize{$\pm$0.34} &64.45\scriptsize{$\pm$0.37} &70.59\scriptsize{$\pm$1.93} &61.57\scriptsize{$\pm$1.75} &66.22\scriptsize{$\pm$3.48}\\
        \cmidrule(lr){1-9}
        Improvement &\textbf{+3.07} &\textbf{+2.11} &\textbf{+5.36} &\textbf{+3.90} &\textbf{+6.29} &\textbf{+3.86} &\textbf{+4.24} &\textbf{+0.43}\\
        
        \midrule
        SSIAT &78.76\scriptsize{$\pm$0.24} &81.64\scriptsize{$\pm$0.34} &77.18\scriptsize{$\pm$0.15} &80.04\scriptsize{$\pm$0.34} &59.57\scriptsize{$\pm$0.32} &66.54\scriptsize{$\pm$1.36} &56.34\scriptsize{$\pm$0.70} &65.62\scriptsize{$\pm$1.43}\\
        w/ Ours &79.13\scriptsize{$\pm$0.22} &82.80\scriptsize{$\pm$0.33} &77.93\scriptsize{$\pm$0.28} &81.68\scriptsize{$\pm$0.27} &63.91\scriptsize{$\pm$0.39} &69.84\scriptsize{$\pm$1.42} &59.97\scriptsize{$\pm$0.57} &68.14\scriptsize{$\pm$1.28}\\
        \cmidrule(lr){1-9}
        Improvement &\textbf{+0.37} &\textbf{+1.16} &\textbf{+0.75} &\textbf{+1.64} &\textbf{+4.34} &\textbf{+3.30} &\textbf{+3.63} &\textbf{+2.52}\\

        \midrule
        MOS &74.07\scriptsize{$\pm$0.36} &78.84\scriptsize{$\pm$0.43} &71.50\scriptsize{$\pm$0.20} &76.92\scriptsize{$\pm$0.22} &57.71\scriptsize{$\pm$0.55} &65.84\scriptsize{$\pm$1.00} &56.06\scriptsize{$\pm$0.08} &65.71\scriptsize{$\pm$1.02}\\
        w/ Ours &77.03\scriptsize{$\pm$0.43} &81.68\scriptsize{$\pm$0.51} &76.54\scriptsize{$\pm$0.37} &80.97\scriptsize{$\pm$0.26} &62.87\scriptsize{$\pm$0.82} &68.91\scriptsize{$\pm$1.68} &61.54\scriptsize{$\pm$0.31} &68.50\scriptsize{$\pm$1.01}\\
        \cmidrule(lr){1-9}
        Improvement &\textbf{+2.96} &\textbf{+2.84} &\textbf{+5.04} &\textbf{+4.05} &\textbf{+5.16} &\textbf{+3.07} &\textbf{+5.48} &\textbf{+2.79}\\
        \bottomrule
    \end{tabular}
    \end{table*}

\section{Experiments}
\label{experiments}

\begin{table*}[ht]
    \caption{Performance comparison (\%) between Aper and ACL.  \textbf{Bolded} indicates the best performance, \underline{underline} denotes the second best. `Improvement' represents the boost of ACL towards the best variants of Aper.}
    \label{tab:result_aper}
    \centering
    \setlength{\tabcolsep}{1.5mm}{
    \begin{tabular}{@{\hspace{2mm}}lcccccccc@{\hspace{2mm}}}
        \toprule
        \multirow{2.4}*{Method} &\multicolumn{2}{c}{ImageNet-R-Inc20} &\multicolumn{2}{c}{ImageNet-R-Inc10} &\multicolumn{2}{c}{ImageNet-A-Inc20} &\multicolumn{2}{c}{ImageNet-A-Inc10}\\
        \cmidrule(lr){2-9}
        &LA &AIA &LA &AIA &LA &AIA &LA &AIA\\
        \midrule
        SimpleCIL &61.35\scriptsize{$\pm$0.00} &66.97\scriptsize{$\pm$0.46} &61.35\scriptsize{$\pm$0.00} &67.58\scriptsize{$\pm$0.47} &49.24\scriptsize{$\pm$0.00} &58.35\scriptsize{$\pm$1.16} &49.24\scriptsize{$\pm$0.00} &59.33\scriptsize{$\pm$1.01}\\ 
        
        \midrule
        Aper w/ Finetune &63.60\scriptsize{$\pm$1.16} &71.77\scriptsize{$\pm$0.91} &64.19\scriptsize{$\pm$1.11} &71.54\scriptsize{$\pm$1.02} &\underline{51.74}\scriptsize{$\pm$1.91} &\underline{60.65}\scriptsize{$\pm$1.94} &\underline{50.44}\scriptsize{$\pm$2.34} &\underline{60.71}\scriptsize{$\pm$2.09}\\

        Aper w/ VPT-Deep &68.70\scriptsize{$\pm$5.76} &75.08\scriptsize{$\pm$6.13} &\underline{68.00}\scriptsize{$\pm$1.05} &\underline{74.71}\scriptsize{$\pm$1.34} &46.11\scriptsize{$\pm$3.25} &56.03\scriptsize{$\pm$3.22} &42.15\scriptsize{$\pm$4.09} &53.01\scriptsize{$\pm$4.91} \\

        Aper w/ VPT-Shallow &64.50\scriptsize{$\pm$0.72} &70.21\scriptsize{$\pm$0.91} &64.83\scriptsize{$\pm$0.38} &71.20\scriptsize{$\pm$0.71} &46.90\scriptsize{$\pm$1.46} &56.42\scriptsize{$\pm$0.83} &45.61\scriptsize{$\pm$1.84} &56.55\scriptsize{$\pm$2.36} \\
        
        Aper w/ SSF &\underline{70.07}\scriptsize{$\pm$0.37} &\underline{76.29}\scriptsize{$\pm$0.80} &67.84\scriptsize{$\pm$0.06} &74.31\scriptsize{$\pm$0.44} &50.24\scriptsize{$\pm$1.47} &59.65\scriptsize{$\pm$0.94} &47.93\scriptsize{$\pm$1.57} &58.59\scriptsize{$\pm$1.16}\\ 

        Aper w/ Adapter &67.25\scriptsize{$\pm$1.21} &73.13\scriptsize{$\pm$1.54} &62.47\scriptsize{$\pm$0.17} &68.70\scriptsize{$\pm$0.66} &49.22\scriptsize{$\pm$0.03} &58.37\scriptsize{$\pm$1.17} &49.23\scriptsize{$\pm$0.07} &59.34\scriptsize{$\pm$1.03}\\
        
        \midrule
        ACL (Ours) &\textbf{73.93}\scriptsize{$\pm$0.38} &\textbf{77.90}\scriptsize{$\pm$0.57} &\textbf{72.26}\scriptsize{$\pm$0.43} &\textbf{76.33}\scriptsize{$\pm$0.46} &\textbf{56.88}\scriptsize{$\pm$0.31} &\textbf{63.50}\scriptsize{$\pm$1.85} &\textbf{55.18}\scriptsize{$\pm$0.34} &\textbf{62.64}\scriptsize{$\pm$1.53}\\
        
        \midrule
        Improvement &\textbf{+3.86} &\textbf{+1.61} &\textbf{+4.26} &\textbf{+1.62} &\textbf{+5.14} &\textbf{+2.85} &\textbf{+4.74} &\textbf{+1.93}\\
        \bottomrule
    \end{tabular}
    }
    \end{table*}

\subsection{Experiment Setup}

\textbf{Dataset.} Given that PTMs are typically trained on ImageNet series datasets~\cite{ridnik2021imagenet}, evaluation on the standard ImageNet benchmark is meaningless due to the overlapping data distribution~\cite{zhou2024revisiting}. Hence, we evaluate ACL on two datasets that exhibit a significant domain gap~\cite{zhou2024revisiting} with ImageNet, \textit{i.e.}, ImageNet-R~\cite{hendrycks2021many} and ImageNet-A~\cite{hendrycks2021natural}. To simulate a CL scenario, both datasets are equally divided into multiple tasks without overlapping data. Specifically, we create two task configurations: (1) 20 tasks with \textbf{10} classes each (Inc-10) and (2) 10 tasks with \textbf{20} classes each (Inc-20). 

\textbf{Baselines.} We compare our proposed method against six state-of-the-art PTM-based CL methods: L2P~\cite{wang2022learning}, DualPrompt~\cite{wang2022dualprompt}, RanPAC~\cite{mcdonnell2024ranpac}, FeCAM~\cite{goswami2024fecam}, SSIAT~\cite{tan2024semantically}, and MOS~\cite{sun2024mos}. Since our framework is designed as a plug-and-play component, we integrate it into these baseline methods to systematically assess its effectiveness. We note that L2P and DualPrompt use a linear classifier, and other methods use a cosine classifier or its variants.
Furthermore, we include a comparison with Aper~\cite{zhou2024revisiting}, a method that fine-tunes the PTM on the first task using standard classification loss and freezes the resulting model for future tasks. Aper is specialized through various adaptation algorithms, including full \textbf{Finetune}, Visual Prompt Tuning (\textbf{VPT}), Scale and Shift (\textbf{SSF}), and \textbf{Adapter}-based tuning. In particular, Aper with VPT has two variants: \textbf{VPT-Deep}, which prepends the prompts at every attention layer, and \textbf{VPT-Shallow}, which only prepends the prompts at the first layer~\cite{zhou2024revisiting}. Following these baselines, our validation focuses on a general and realistic CL scenario \textit{i.e.}, class incremental learning~\cite{wang2023comprehensive}. 

\textbf{Implementation Details.} We select a representative PTM, denoted as ViT-B/16-IN1K, for our experiments. This PTM is initially pre-trained on ImageNet21K~\cite{ridnik2021imagenet} and subsequently finetuned on ImageNet1K~\cite{deng2009imagenet}. To ensure consistency and reproducibility, we adhere to the hyperparameter configurations provided by the open-source library PILOT~\cite{sun2023pilot} for all CL methods. For each incremental task, we limit the adaptation phase to 1 training epoch to minimize computational overhead and set $\tau=0.1$ for ACL loss. For all results, we report mean $\pm$ std of 5 runs with different task orders.

\textbf{Evaluation Metrics.} In line with established conventions~\cite{zhou2024continual, wang2023comprehensive}, the CL performance is evaluated using two key metrics: {\em Last Accuracy} (LA) and {\em Average Incremental Accuracy} (AIA). LA measures the model's performance across all classes after completing the final task, while AIA quantifies the average performance of the model after learning each incremental task~\cite{lu2024revisiting}. Formally, let $K$ denote the total number of tasks, and let $A_{b}$ represent the classification accuracy evaluated on the test set encompassing all classes learned up to and including the $b$-th task. These metrics are defined as $LA = A_{K}$ and $AIA = \frac{1}{K} \sum_{b=1}^{K} A_{b}$. \textit{Higher values for LA and AIA indicate superior CL performance, reflecting a better balance between stability and plasticity.}

\subsection{Main Results}
\label{main_results}

\textbf{Integration with Existing Methods.} We begin by assessing ACL's effectiveness when incorporated into six state-of-the-art PTM-based CL methods. As shown in Tab.~\ref{tab:result_integration}, ACL consistently improves the performance of these methods across diverse datasets and incremental steps. Notably, ACL achieves gains of up to 7.87\% in LA and 5.98\% in AIA compared to the original methods. These findings highlight the effectiveness of ACL in achieving a better stability-plasticity trade-off, thereby enhancing the performance of existing CL methods.

\textbf{Comparison with Aper.} We further compare ACL with Aper, a method that adapts the PTM using a standard classification loss on the first task only. To ensure a fair comparison, we integrate ACL with SimpleCIL~\cite{zhou2024revisiting}, which uses the same cosine classifier as Aper and involves no training after adaptation. As shown in Tab.~\ref{tab:result_aper}, ACL consistently surpasses Aper across all datasets and incremental steps. Notably, ACL achieves performance gains of up to 5.14\% in LA and 2.85\% in AIA compared to the best-performing variants of Aper. These results demonstrate ACL’s superiority in leading to a better stability-plasticity balance than Aper.

\subsection{Ablation Study}

\begin{figure}[t]
    \centering
    {\includegraphics[width = 1\linewidth]{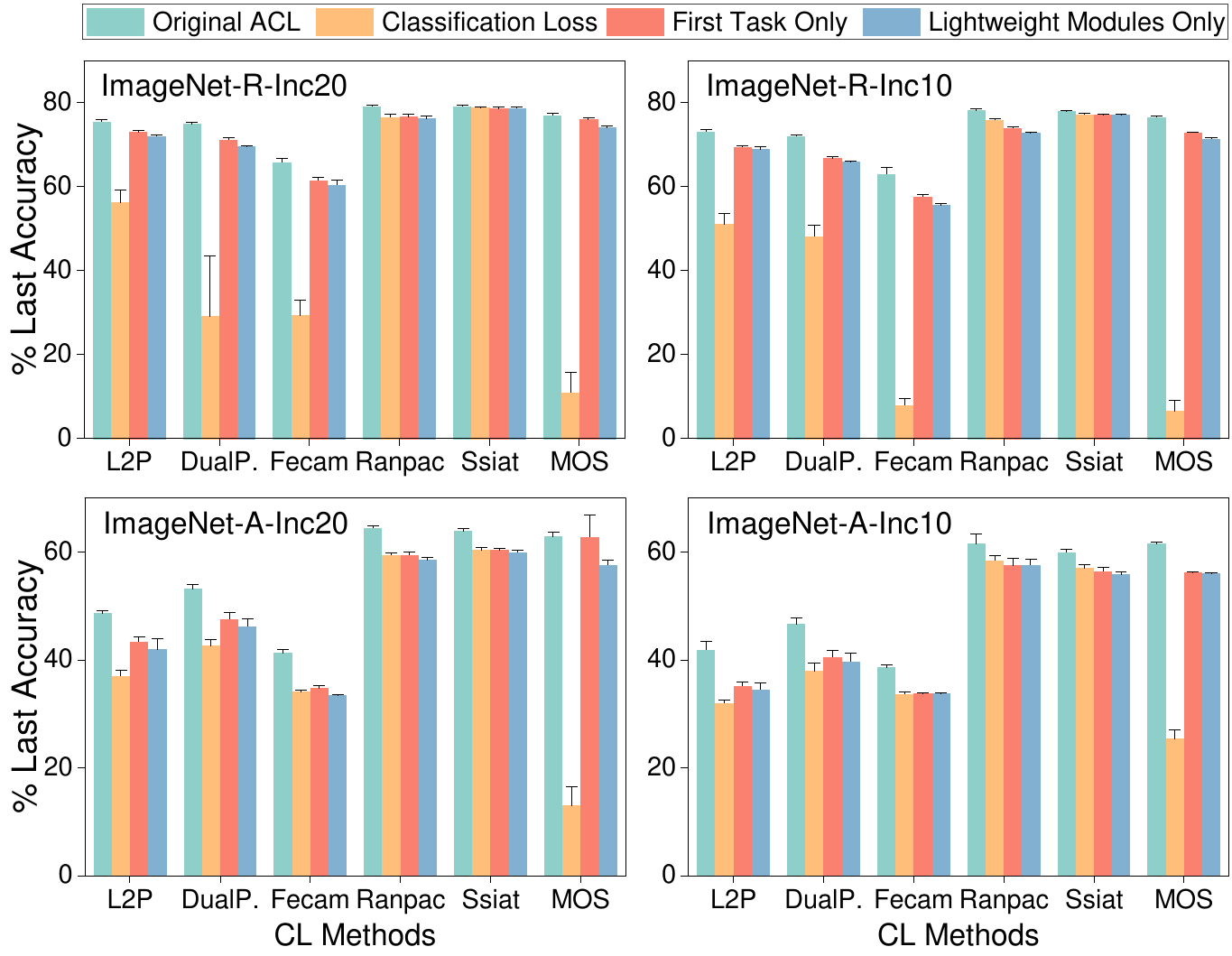}}
    \caption{Performance of original ACL and its ablation variants, including (1) using standard \textbf{classification loss} for adaptation, (2) adapting for the \textbf{first task only}, and (3) adapting \textbf{lightweight modules only} with frozen backbone.
}
\label{fig:result_ab}
\end{figure}

\textbf{Key Components for Adaptation.} We first conduct an ablation study to systematically investigate the impact of various strategies employed during the adaptation phase, focusing on three key aspects: loss design, adaptation steps, and adapted network components. As illustrated in Fig.~\ref{fig:result_ab}, all ablation variants exhibit inferior performance compared to the original ACL framework across all integrated CL methods. These results underscore the critical importance of \textit{continually adapting the entire PTM backbone using the proposed ACL loss for all incremental tasks}, which facilitates more effective adaptation and knowledge retention.

\begin{figure}[ht]
    \centering
    \subcaptionbox{RanPAC}{\includegraphics[width = 0.48\linewidth]{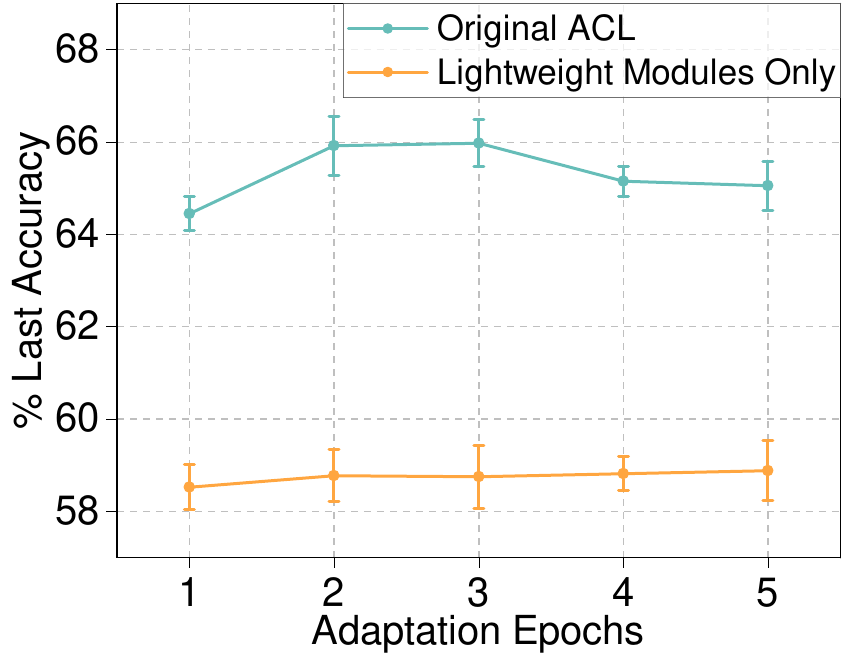}}
    \hfill
    \subcaptionbox{SSIAT}{\includegraphics[width = 0.48\linewidth]{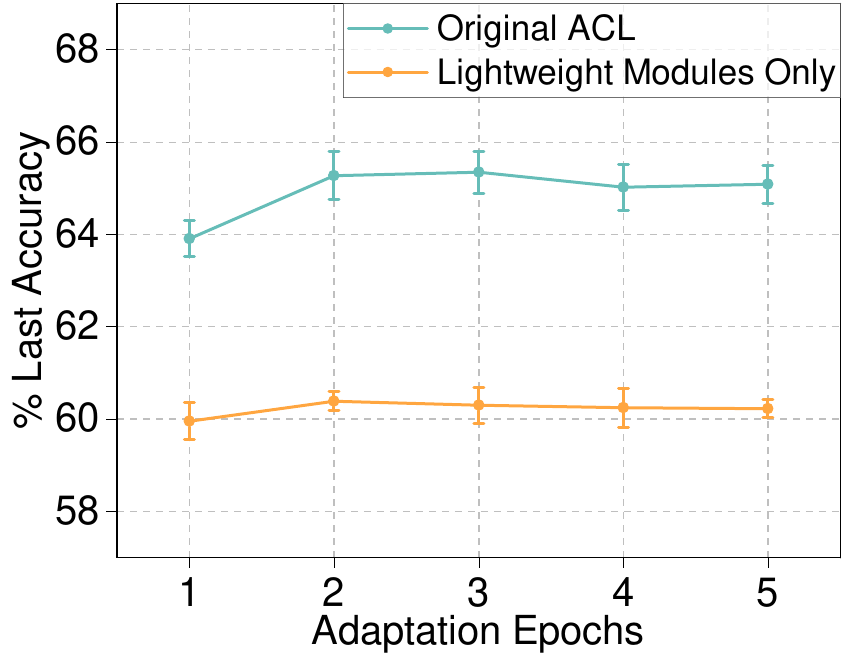}}
    \caption{Performance with different adaptation epochs.
}
\label{fig:multi_epoch}
\end{figure}

\textbf{Full PTM Adaptation vs. Multi-Epoch Adaptation.}
To further show the advantages of adapting the entire PTM versus solely adapting lightweight modules, we extended our comparison by considering the impact of multiple adaptation epochs. The results, presented in Fig.~\ref{fig:multi_epoch}, demonstrate two key findings: (1) adapting the entire PTM consistently outperforms adapting only lightweight modules; (2) increasing adaptation epochs beyond two yields marginal or negligible performance gains for either strategy. These observations indicate that the performance benefits derived from full PTM adaptation cannot be replicated merely by increasing the adaptation epochs. This reinforces our claim that freezing pre-trained weights results in a suboptimal balance between stability and plasticity.

\subsection{Visualization} 

\begin{figure}[ht]
    \centering
    \subcaptionbox{Frozen}{\includegraphics[width = 0.48\linewidth]{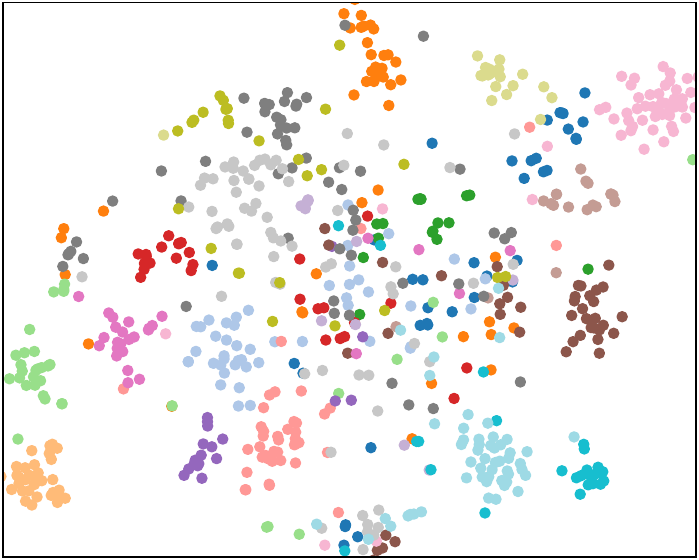}}
    \hfill
    \subcaptionbox{w/ ACL (Ours)}{\includegraphics[width = 0.48\linewidth]{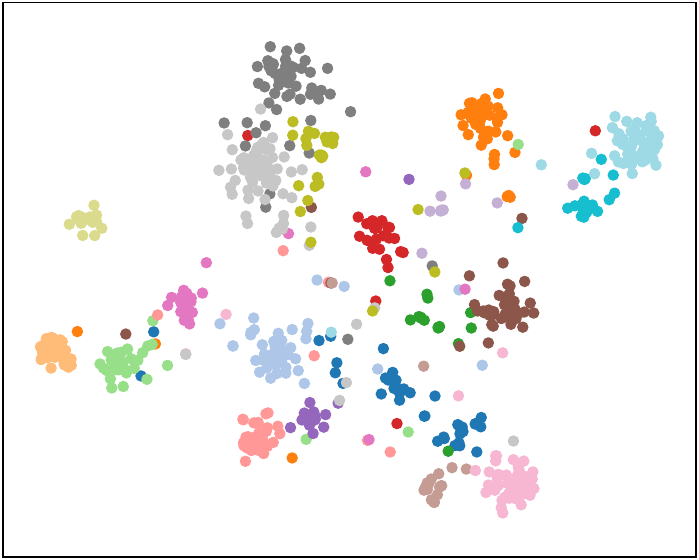}}
    \\
    \subcaptionbox{Grad-CAM}{\includegraphics[width = 1\linewidth]{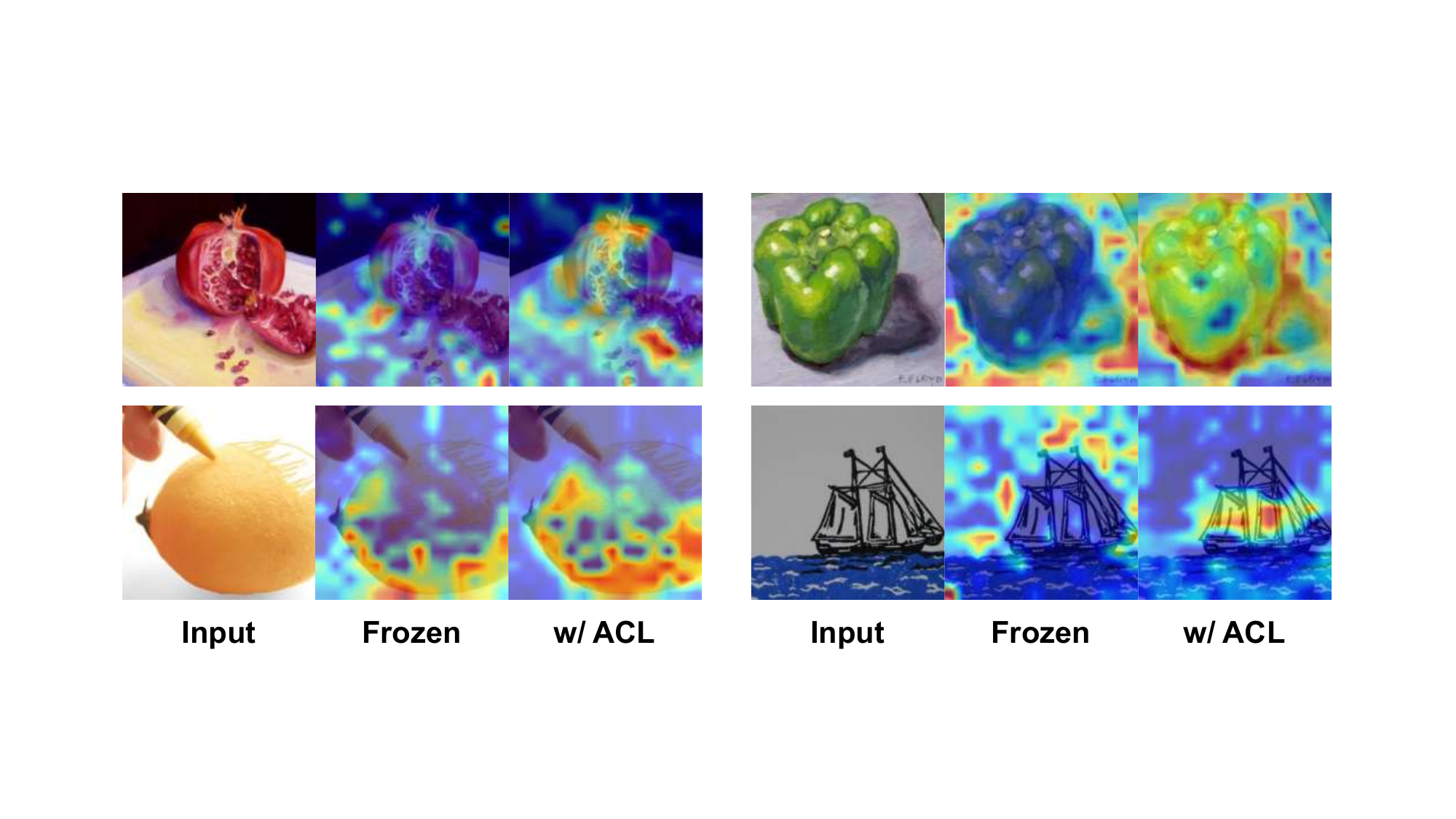}}
    \caption{(a-b) Visualization of 2D feature representations using t-SNE. (c) Grad-CAM visualization, where important regions are highlighted with warm colors.
}
\label{fig:visualization}
\end{figure}

We employ t-SNE~\cite{van2008visualizing} to visualize the feature representations extracted by the final PTMs, comparing those obtained without and with the proposed ACL framework. For simplicity, we adopt SimpleCIL~\cite{zhou2024revisiting} as the baseline method and evaluate on ImageNet-R-Inc20, selecting two classes per incremental task for clearer visualization. The results, shown in Fig.~\ref{fig:visualization}(a) and (b), demonstrate that the PTM adapted with ACL generates more discriminative feature representations than the frozen model, even for classes in previously learned tasks. This indicates that ACL effectively enhances feature discriminability across all incremental tasks, achieving a better stability-plasticity trade-off.

To further validate our approach, we visualize Grad-CAM~\cite{selvaraju2017grad} results on samples with a large domain gap~\cite{hendrycks2021many} relative to the pre-training data, which highlight critical image regions for concept prediction. As depicted in Fig.~\ref{fig:visualization}(c), the frozen PTM often attends to irrelevant background regions. In contrast, the PTM adapted via ACL focuses more accurately on class-specific features. These findings underscore the necessity of adapting PTMs to incremental data, especially when the target distribution significantly diverges from the pre-training domain.

\subsection{Validation on Other Backbones}

\textbf{Validation on ViT-B/16-IN21K.}
To further validate the effectiveness of ACL, we conduct experiments on ViT-B/16-IN21K, a model pre-trained on ImageNet21K only, with ImageNet-A-Inc20 as the benchmark. As shown in Tab.~\ref{tab:result_in21k}, ACL consistently enhances the CL performance across various CL methods. These findings underscore the versatility and generalizability of our framework.

\begin{table}[ht]
\caption{Performance (\%LA) using ViT-B/16-IN21K. `Improv.' represents the boost of ACL towards original methods.}
\label{tab:result_in21k}
\centering
\begin{tabular}{@{\hspace{2mm}}lccc@{\hspace{2mm}}}
\toprule
Methods & Original & w/ ACL (Ours) & Improv. \\
\midrule
L2P & 39.83\scriptsize{$\pm$1.15} & 45.49\scriptsize{$\pm$0.56} &\textbf{+5.66} \\
DualPrompt & 43.02\scriptsize{$\pm$1.29} & 45.57\scriptsize{$\pm$1.56} &\textbf{+2.55} \\
FeCAM & 45.37\scriptsize{$\pm$0.30} & 47.81\scriptsize{$\pm$1.22} &\textbf{+2.44} \\
RanPAC & 54.59\scriptsize{$\pm$0.84} & 58.97\scriptsize{$\pm$0.21} &\textbf{+4.38} \\
SSIAT & 56.84\scriptsize{$\pm$0.49} & 59.80\scriptsize{$\pm$0.33} &\textbf{+2.96} \\
MOS & 54.17\scriptsize{$\pm$0.45} & 58.55\scriptsize{$\pm$0.24} &\textbf{+4.38} \\
\bottomrule
\end{tabular}
\end{table}

\textbf{Validation on CLIP.}
While our study primarily focuses on visual models, the insights presented in our paper are potentially applicable to visual-language models, such as CLIP~\cite{radford2021learning}. To demonstrate this, we employ Continual CLIP~\cite{thengane2022continualclip} as the baseline and evaluate ACL on the ImageNet-R-inc20 benchmark. Since the text labels for the same class are consistent, we only adapt the visual encoder using ACL. The experimental results, summarized in Tab.~\ref{tab:result_clip}, indicate that ACL significantly enhances the CL performance of CLIP. These findings demonstrate the potential of ACL to improve CL in the context of visual-language models.

\begin{table}[ht]
    \caption{Performance (\%) using CLIP with/without ACL.}
    \label{tab:result_clip}
    \centering
    \setlength{\tabcolsep}{1.5mm}{
    \begin{tabular}{@{\hspace{2mm}}lcc@{\hspace{2mm}}}
        \toprule
        Method &LA &AIA\\
        \midrule
        Continual CLIP &71.70\scriptsize{$\pm$0.01} &78.73\scriptsize{$\pm$0.66} \\ 
        w/ Ours &\textbf{74.98}\scriptsize{$\pm$0.25} \normalsize(\textbf{+3.28}) &\textbf{80.95}\scriptsize{$\pm$0.51} \normalsize(\textbf{+2.22}) \\ 
        \bottomrule
    \end{tabular}
    }
    \end{table}

\section{Conclusion}

In this paper, we revisit CL with PTMs and argue that existing PTM-based CL methods overly prioritize stability at the expense of plasticity. To address this limitation, we propose ACL, a framework that can be orthogonally integrated with existing PTM-based CL methods to enhance plasticity while simultaneously maintaining stability. Extensive experiments demonstrate the effectiveness of ACL in enhancing plasticity and achieving a more balanced stability-plasticity trade-off. Future work will focus on exploring more effective or efficient adaptation algorithms within the ACL framework to further improve its performance and applicability.

\textbf{Limitations.} Adapting the entire PTM introduces additional GPU memory consumption, specifically, approximately 7GB with the experiment settings in Table~\ref{tab:result_integration}.

\section{Appendix}

\subsection{Proofs of Lemmas}
\setcounter{lemma}{0}
In this section, we present the proofs for the two lemmas used in the main text.

\begin{lemma}
For any two $\ell_2$-normalized vectors $a, b \in \mathbb{S}^{d-1}$, the squared Euclidean distance is directly related to their cosine similarity:
\begin{equation}
\|a - b\|_2^2 = 2\bigl(1 - \cos(a, b)\bigr).
\end{equation}
\end{lemma}

\begin{proof}
Since $a, b \in \mathbb{S}^{d-1}$ be two $\ell_2$-normalized vectors, $\|a\|_2 = \|b\|_2 = 1$. We expand the squared Euclidean distance between $a$ and $b$:
\begin{equation}
\begin{aligned}
\|a - b\|_2^2 &= (a - b)^\top(a - b) \\
&= a^\top a - 2a^\top b + b^\top b \\
&= \|a\|_2^2 + \|b\|_2^2 - 2a^\top b \\
&= 1 + 1 - 2a^\top b \\
&= 2(1 - a^\top b).
\end{aligned}
\end{equation}

Since both vectors are normalized, we have $\cos(a, b) = \frac{a^\top b}{\|a\|_2\|b\|_2} = a^\top b$. Therefore,
\begin{equation}
\|a - b\|_2^2 = 2(1 - \cos(a, b)).
\end{equation}
\end{proof}

\begin{lemma}
For a give class $y_{i}$, the prototype is the unique point that minimizes the expected squared Euclidean distance to all features.
\begin{equation}
p_{y_{i}}=\underset{z}{\arg \min } \mathbb{E}_{(x, y) \in \mathcal{D}_{k, y=y_{i}}}\left[\|\phi(x)-z\|_{2}^{2}\right].
\end{equation}
\end{lemma}

\begin{proof}
Consider the optimization problem:
\begin{equation}
u_{y_i} = \arg\min_{z} \mathbb{E}_{(x, y) \in \mathcal{D}_{k, y=y_i}}\left[\|\phi(x) - z\|_2^2\right].
\end{equation}

To find the minimizer, we compute the gradient with respect to $z$:
\begin{equation}
    \begin{aligned}
    &\nabla_z \mathbb{E}_{(x, y) \in \mathcal{D}_{k, y=y_i}}\left[\|\phi(x) - z\|_2^2\right] \\
    &= \nabla_z \mathbb{E}_{(x, y) \in \mathcal{D}_{k, y=y_i}}\left[(\phi(x) - z)^\top(\phi(x) - z)\right] \\
    &= \mathbb{E}_{(x, y) \in \mathcal{D}_{k, y=y_i}}\left[-2(\phi(x) - z)\right] \\
    &= -2\left(\mathbb{E}_{(x, y) \in \mathcal{D}_{k, y=y_i}}[\phi(x)] - z\right).
    \end{aligned}
\end{equation}

Setting the gradient to zero yields:
\begin{equation}
\mathbb{E}_{(x, y) \in \mathcal{D}_{k, y=y_i}}[\phi(x)] - z = 0,
\end{equation}
which implies
\begin{equation}
z = \mathbb{E}_{(x, y) \in \mathcal{D}_{k, y=y_i}}[\phi(x)].
\end{equation}

The second-order condition confirms this is a minimum since the Hessian is $2I$, which is positive definite. Therefore, the prototype $p_{y_i}$ is uniquely given by:
\begin{equation}
p_{y_i} = \mathbb{E}_{(x, y) \in \mathcal{D}_{k, y=y_i}}[\phi(x)].
\end{equation}
\end{proof}

\subsection{Implementation Details}
\textbf{Training Settings.} During the adaptation phase, we use a learning rate of 1e-6 for L2P and DualPrompt, and 1e-4 for all other methods. Following established conventions~\cite{wang2022learning}, all models are trained using a batch size of 128. For data splitting and preprocessing, we follow the open-source library PILOT~\cite{sun2023pilot}. All experiments are repeated with five random seeds: 1993, 1994, 1995, 1996, and 1997.

\textbf{Hardware Configuration.} All experiments are conducted on RTX 3090 GPUs, with each experiment fitting within a single 24GB GPU.

\subsection{Impact of Temperature Settings.} 

We investigate the impact of the temperature parameter ($\tau$) in the ACL loss on the CL performance. As shown in Table~\ref{tab:result_temp}, ACL consistently improves performance across various CL methods over a wide range of temperature values.

\begin{table*}[ht]
    \caption{Performance (\%LA) of ACL across temperature ($\tau$) settings.}
    \label{tab:result_temp}
    \centering
    \begin{tabular}{@{\hspace{2mm}}lcccccc@{\hspace{2mm}}}
    \toprule
    Method & without ACL & $\tau$=0.02  & $\tau$=0.05  & $\tau$=0.1 (current) & $\tau$=0.2   & $\tau$=0.5   \\
    \midrule
    L2P     & 42.58\scriptsize{$\pm$0.39} & 48.20\scriptsize{$\pm$1.33} & 49.14\scriptsize{$\pm$0.66} & 48.65\scriptsize{$\pm$0.55} & 47.18\scriptsize{$\pm$0.71} & 45.75\scriptsize{$\pm$1.08} \\
    DualPrompt     & 45.35\scriptsize{$\pm$1.04} & 50.75\scriptsize{$\pm$0.47} & 51.94\scriptsize{$\pm$0.45} & 53.22\scriptsize{$\pm$0.70} & 53.04\scriptsize{$\pm$0.88} & 52.60\scriptsize{$\pm$0.89} \\
    Fecam     & 33.43\scriptsize{$\pm$0.18} & 45.57\scriptsize{$\pm$1.65} & 41.41\scriptsize{$\pm$0.65} & 41.28\scriptsize{$\pm$0.61} & 39.50\scriptsize{$\pm$0.36} & 36.39\scriptsize{$\pm$0.32} \\
    Ranpac     & 58.16\scriptsize{$\pm$0.46} & 62.87\scriptsize{$\pm$1.06} & 63.74\scriptsize{$\pm$0.46} & 64.45\scriptsize{$\pm$0.37} & 64.26\scriptsize{$\pm$0.57} & 62.04\scriptsize{$\pm$0.39} \\
    SSIAT      & 59.57\scriptsize{$\pm$0.32} & 63.00\scriptsize{$\pm$0.28} & 63.86\scriptsize{$\pm$0.49} & 63.91\scriptsize{$\pm$0.39} & 63.50\scriptsize{$\pm$0.41} & 62.58\scriptsize{$\pm$0.17} \\
    MOS        & 57.71\scriptsize{$\pm$0.55} & 60.98\scriptsize{$\pm$1.15} & 62.71\scriptsize{$\pm$0.52} & 62.87\scriptsize{$\pm$0.82} & 62.98\scriptsize{$\pm$0.80} & 61.25\scriptsize{$\pm$0.41} \\
    \bottomrule
    \end{tabular}
\end{table*}

\subsection{Full PTM Adaptation vs. Multi-Epoch Adaptation}
\begin{figure}[ht]
    \centering
    \subcaptionbox{L2P}{\includegraphics[width = 0.48\linewidth]{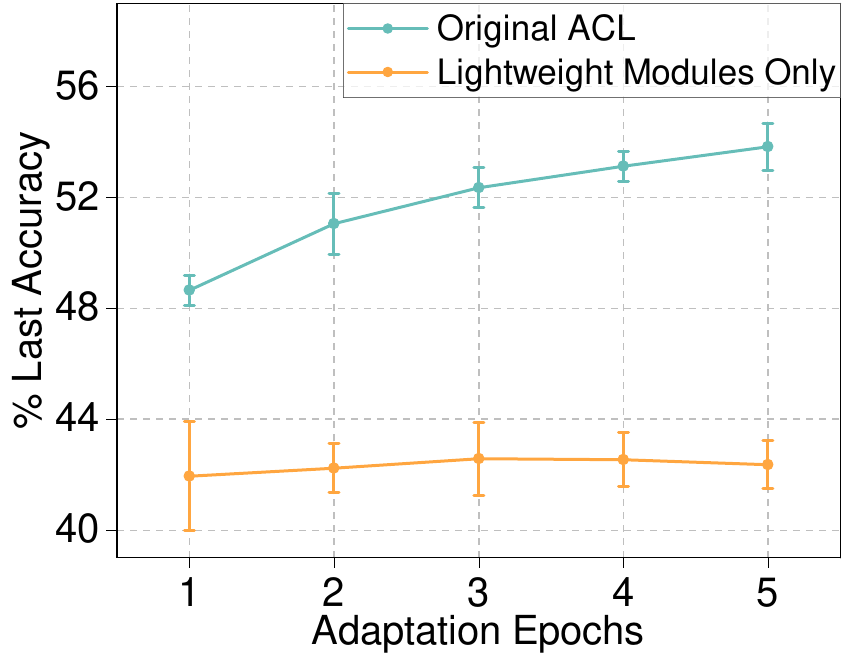}}
    \hfill
    \subcaptionbox{DualPrompt}{\includegraphics[width = 0.48\linewidth]{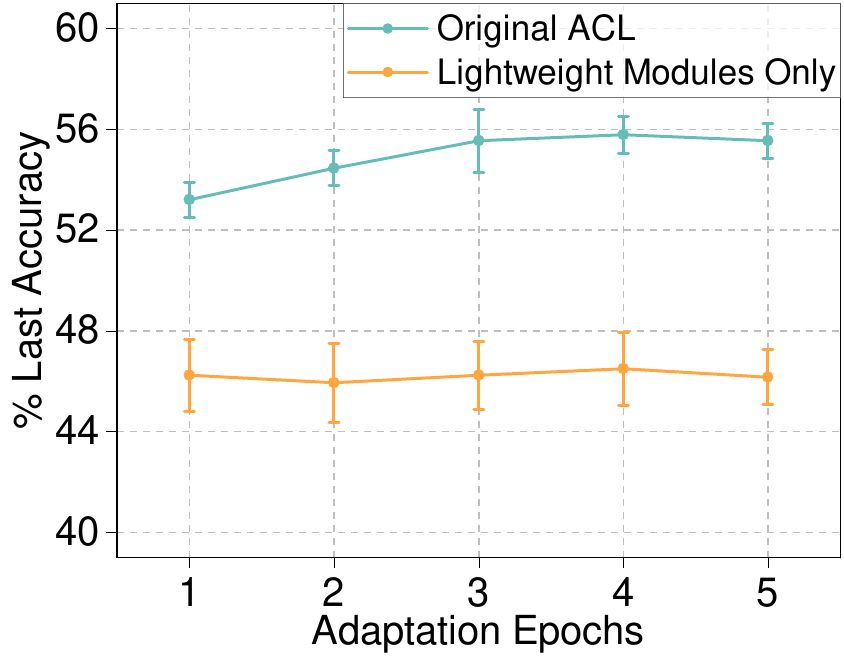}}
    \\
    \subcaptionbox{Fecam}{\includegraphics[width = 0.48\linewidth]{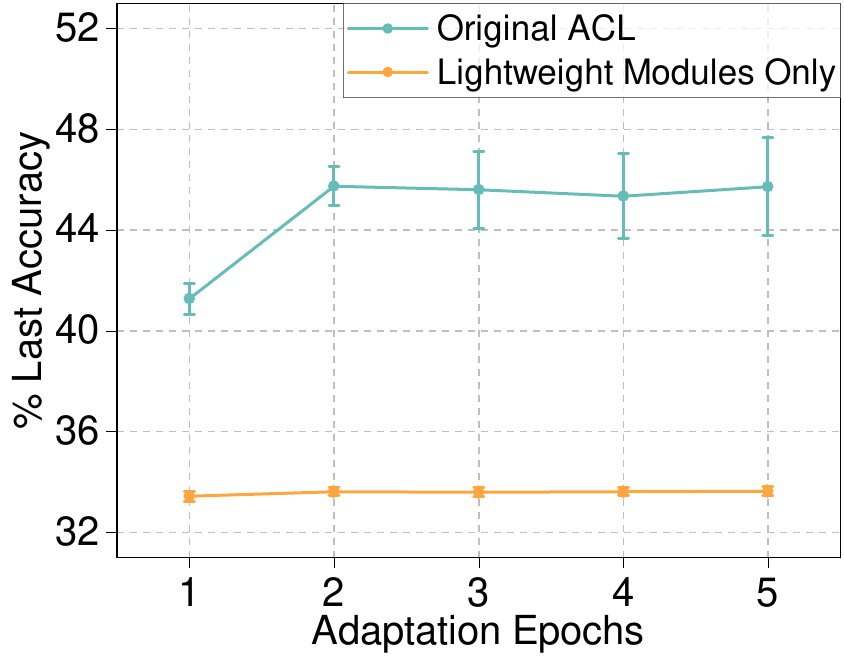}}
    \hfill
    \subcaptionbox{MOS}{\includegraphics[width = 0.48\linewidth]{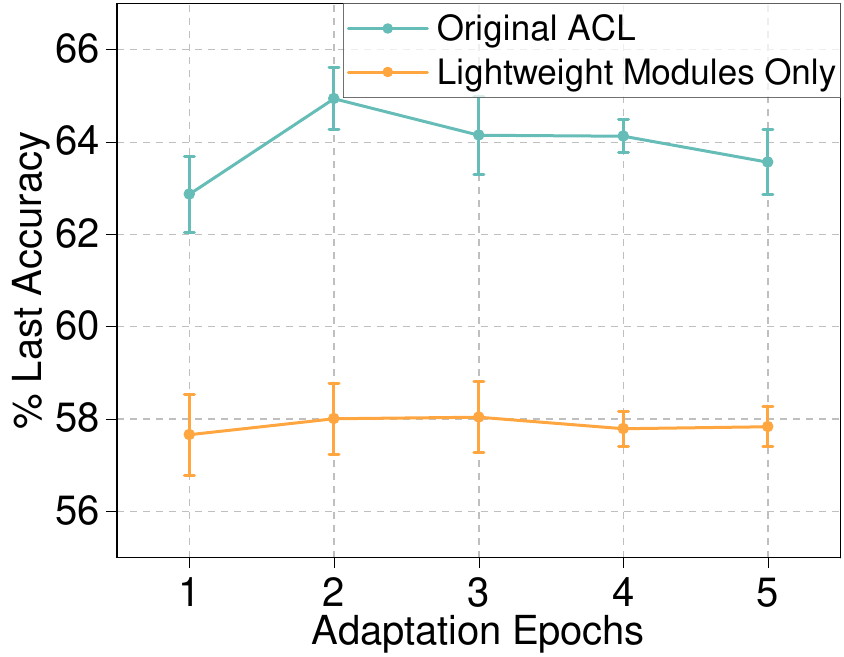}}
    \caption{Performance with different adaptation epochs.}
\label{fig:multi_epoch_2}
\end{figure}

The additional results of comparison between full PTM adaptation and multi-epoch adaptation are presented in Fig.~\ref{fig:multi_epoch_2}. The results indicate that simply increasing the number of adaptation epochs cannot replicate the performance gains achieved through full PTM adaptation.

\subsection{Benchmark Selection Principle}

\begin{table*}[ht]
    \caption{Performance of ViT-B/16-IN21K on multiple datasets with SimpleCIL. All results are sourced from ~\cite{zhou2024revisiting}.}
    \label{tab:result_zs}
    \centering
    \begin{tabular}{@{\hspace{2mm}}lccccccc@{\hspace{2mm}}}
        \toprule
        Dataset &CIFAR100 &CUB &OmniBench &VTAB &ObjectNet &ImageNet-R	&ImageNet-A\\
        \midrule
        LA (\%) &81.26 &86.73 &73.15 &84.38 &53.59 &54.55 &49.44 \\ 
        \bottomrule
    \end{tabular}
\end{table*}

This work focuses on CL scenarios where a significant domain gap exists between the pre-trained dataset (e.g., ImageNet-1K/21K) and downstream tasks. Such settings are common in real-world applications and pose substantial challenges to model plasticity. Datasets like ImageNet-R and ImageNet-A exemplify these large domain shifts.

In contrast, some commonly used datasets in prior CL research (e.g., CIFAR100~\cite{krizhevsky2009learning}, CUB~\cite{wah2011caltech}, OmniBench~\cite{zhang2022benchmarking} and VTAB~\cite{zhai2019large}) exhibit relatively small distributional gaps with ImageNet, which does not align with our focus. To illustrate this, Table~\ref{tab:result_zs} reports the zero-shot performance of ViT-B/16-IN21K on multiple benchmarks using SimpleCIL~\cite{zhou2024revisiting}. High accuracy on datasets like CIFAR100 (81.26\%), CUB (86.73\%), OmniBench (73.15\%), and VTAB (84.38\%) indicates a small domain shift from ImageNet. In contrast, ObjectNet (53.59\%), ImageNet-R (54.55\%), and ImageNet-A (49.44\%) show significantly lower accuracy, confirming their suitability for evaluation under large domain gaps.

\subsection{Validation on ObjectNet}
Given its significant domain divergence from ImageNet (as demonstrated in Table~\ref{tab:result_zs}), ObjectNet~\cite{barbu2019objectnet} also serves as an appropriate benchmark for validating our method's robustness. We evaluate our approach on ObjectNet-inc20 using the data preprocessing protocol from~\cite{zhou2024revisiting}. Table~\ref{tab:result_obj} presents the performance of several CL methods before and after integrating our ACL framework. Results show that ACL generally enhances performance across most methods in this challenging setting.

\begin{table}[ht]
    \caption{Performance (\%LA) on ObjectNet.}
    \label{tab:result_obj}
    \centering
    \begin{tabular}{@{\hspace{2mm}}lcccccc@{\hspace{2mm}}}
        \toprule
        & Original & w/ Ours & Improvement \\
        \midrule
        L2P & 55.91$\pm$\scriptsize{0.33} & 58.49$\pm$\scriptsize{0.73} & \textbf{+2.58} \\
        DualP. & 53.99$\pm$\scriptsize{0.30} & 57.19$\pm$\scriptsize{0.20} & \textbf{+3.20} \\
        FeCAM & 54.38$\pm$\scriptsize{0.57} & 56.57$\pm$\scriptsize{0.58} & \textbf{+2.19} \\
        RanPAC & 63.79$\pm$\scriptsize{0.12} & 64.92$\pm$\scriptsize{0.29} & \textbf{+1.13} \\
        SSIAT & 64.63$\pm$\scriptsize{0.28} & 65.22$\pm$\scriptsize{0.26} & \textbf{+0.59} \\
        MOS & 62.75$\pm$\scriptsize{0.30} & 60.06$\pm$\scriptsize{1.36} & -2.69 \\
        \bottomrule
    \end{tabular}
\end{table}

\bibliography{ref}

\begin{thebibliography}{57}
\providecommand{\natexlab}[1]{#1}

\bibitem[{Aljundi et~al.(2019)Aljundi, Lin, Goujaud, and Bengio}]{aljundi2019gradient}
Aljundi, R.; Lin, M.; Goujaud, B.; and Bengio, Y. 2019.
\newblock Gradient based sample selection for online continual learning.
\newblock \emph{Advances in neural information processing systems}, 32.

\bibitem[{Barbu et~al.(2019)Barbu, Mayo, Alverio, Luo, Wang, Gutfreund, Tenenbaum, and Katz}]{barbu2019objectnet}
Barbu, A.; Mayo, D.; Alverio, J.; Luo, W.; Wang, C.; Gutfreund, D.; Tenenbaum, J.; and Katz, B. 2019.
\newblock Objectnet: A large-scale bias-controlled dataset for pushing the limits of object recognition models.
\newblock \emph{Advances in neural information processing systems}, 32.

\bibitem[{Cha, Lee, and Shin(2021)}]{cha2021co2l}
Cha, H.; Lee, J.; and Shin, J. 2021.
\newblock Co2l: Contrastive continual learning.
\newblock In \emph{Proceedings of the IEEE/CVF International conference on computer vision}, 9516--9525.

\bibitem[{Deng et~al.(2009)Deng, Dong, Socher, Li, Li, and Fei-Fei}]{deng2009imagenet}
Deng, J.; Dong, W.; Socher, R.; Li, L.-J.; Li, K.; and Fei-Fei, L. 2009.
\newblock Imagenet: A large-scale hierarchical image database.
\newblock In \emph{2009 IEEE conference on computer vision and pattern recognition}, 248--255. Ieee.

\bibitem[{Dosovitskiy(2020)}]{dosovitskiy2020image}
Dosovitskiy, A. 2020.
\newblock An image is worth 16x16 words: Transformers for image recognition at scale.
\newblock \emph{arXiv preprint arXiv:2010.11929}.

\bibitem[{Feng, Wang, and Yuan(2022)}]{feng2022overcoming}
Feng, T.; Wang, M.; and Yuan, H. 2022.
\newblock Overcoming catastrophic forgetting in incremental object detection via elastic response distillation.
\newblock In \emph{Proceedings of the IEEE/CVF Conference on Computer Vision and Pattern Recognition}, 9427--9436.

\bibitem[{Gao et~al.(2024)Gao, Dong, He, Wang, and Gong}]{gao2024beyond}
Gao, X.; Dong, S.; He, Y.; Wang, Q.; and Gong, Y. 2024.
\newblock Beyond prompt learning: Continual adapter for efficient rehearsal-free continual learning.
\newblock In \emph{European Conference on Computer Vision}, 89--106. Springer.

\bibitem[{Goodfellow et~al.(2013)Goodfellow, Mirza, Xiao, Courville, and Bengio}]{goodfellow2013empirical}
Goodfellow, I.~J.; Mirza, M.; Xiao, D.; Courville, A.; and Bengio, Y. 2013.
\newblock An empirical investigation of catastrophic forgetting in gradient-based neural networks.
\newblock \emph{arXiv preprint arXiv:1312.6211}.

\bibitem[{Goswami et~al.(2024)Goswami, Liu, Twardowski, and van~de Weijer}]{goswami2024fecam}
Goswami, D.; Liu, Y.; Twardowski, B.; and van~de Weijer, J. 2024.
\newblock Fecam: Exploiting the heterogeneity of class distributions in exemplar-free continual learning.
\newblock \emph{Advances in Neural Information Processing Systems}, 36.

\bibitem[{Grossberg(2013)}]{grossberg2013adaptive}
Grossberg, S. 2013.
\newblock Adaptive Resonance Theory: How a brain learns to consciously attend, learn, and recognize a changing world.
\newblock \emph{Neural networks}, 37: 1--47.

\bibitem[{Hendrycks et~al.(2021{\natexlab{a}})Hendrycks, Basart, Mu, Kadavath, Wang, Dorundo, Desai, Zhu, Parajuli, Guo et~al.}]{hendrycks2021many}
Hendrycks, D.; Basart, S.; Mu, N.; Kadavath, S.; Wang, F.; Dorundo, E.; Desai, R.; Zhu, T.; Parajuli, S.; Guo, M.; et~al. 2021{\natexlab{a}}.
\newblock The many faces of robustness: A critical analysis of out-of-distribution generalization.
\newblock In \emph{Proceedings of the IEEE/CVF international conference on computer vision}, 8340--8349.

\bibitem[{Hendrycks et~al.(2021{\natexlab{b}})Hendrycks, Zhao, Basart, Steinhardt, and Song}]{hendrycks2021natural}
Hendrycks, D.; Zhao, K.; Basart, S.; Steinhardt, J.; and Song, D. 2021{\natexlab{b}}.
\newblock Natural adversarial examples.
\newblock In \emph{Proceedings of the IEEE/CVF conference on computer vision and pattern recognition}, 15262--15271.

\bibitem[{Iscen et~al.(2020)Iscen, Zhang, Lazebnik, and Schmid}]{iscen2020memory}
Iscen, A.; Zhang, J.; Lazebnik, S.; and Schmid, C. 2020.
\newblock Memory-efficient incremental learning through feature adaptation.
\newblock In \emph{Computer Vision--ECCV 2020: 16th European Conference, Glasgow, UK, August 23--28, 2020, Proceedings, Part XVI 16}, 699--715. Springer.

\bibitem[{Jia et~al.(2022)Jia, Tang, Chen, Cardie, Belongie, Hariharan, and Lim}]{jia2022visual}
Jia, M.; Tang, L.; Chen, B.-C.; Cardie, C.; Belongie, S.; Hariharan, B.; and Lim, S.-N. 2022.
\newblock Visual prompt tuning.
\newblock In \emph{European Conference on Computer Vision}, 709--727. Springer.

\bibitem[{Jung et~al.(2023)Jung, Han, Bang, and Song}]{jung2023generating}
Jung, D.; Han, D.; Bang, J.; and Song, H. 2023.
\newblock Generating instance-level prompts for rehearsal-free continual learning.
\newblock In \emph{Proceedings of the IEEE/CVF International Conference on Computer Vision}, 11847--11857.

\bibitem[{Kang et~al.(2022)Kang, Mina, Madjid, Yoon, Hasegawa-Johnson, Hwang, and Yoo}]{kang2022forget}
Kang, H.; Mina, R. J.~L.; Madjid, S. R.~H.; Yoon, J.; Hasegawa-Johnson, M.; Hwang, S.~J.; and Yoo, C.~D. 2022.
\newblock Forget-free continual learning with winning subnetworks.
\newblock In \emph{International Conference on Machine Learning}, 10734--10750. PMLR.

\bibitem[{Khosla et~al.(2020)Khosla, Teterwak, Wang, Sarna, Tian, Isola, Maschinot, Liu, and Krishnan}]{khosla2020supervised}
Khosla, P.; Teterwak, P.; Wang, C.; Sarna, A.; Tian, Y.; Isola, P.; Maschinot, A.; Liu, C.; and Krishnan, D. 2020.
\newblock Supervised contrastive learning.
\newblock \emph{Advances in neural information processing systems}, 33: 18661--18673.

\bibitem[{Kirkpatrick et~al.(2017)Kirkpatrick, Pascanu, Rabinowitz, Veness, Desjardins, Rusu, Milan, Quan, Ramalho, Grabska-Barwinska et~al.}]{kirkpatrick2017overcoming}
Kirkpatrick, J.; Pascanu, R.; Rabinowitz, N.; Veness, J.; Desjardins, G.; Rusu, A.~A.; Milan, K.; Quan, J.; Ramalho, T.; Grabska-Barwinska, A.; et~al. 2017.
\newblock Overcoming catastrophic forgetting in neural networks.
\newblock \emph{Proceedings of the national academy of sciences}, 114(13): 3521--3526.

\bibitem[{Konishi et~al.(2023)Konishi, Kurokawa, Ono, Ke, Kim, and Liu}]{konishi2023parameter}
Konishi, T.; Kurokawa, M.; Ono, C.; Ke, Z.; Kim, G.; and Liu, B. 2023.
\newblock Parameter-level soft-masking for continual learning.
\newblock In \emph{International Conference on Machine Learning}, 17492--17505. PMLR.

\bibitem[{Krizhevsky, Hinton et~al.(2009)}]{krizhevsky2009learning}
Krizhevsky, A.; Hinton, G.; et~al. 2009.
\newblock Learning multiple layers of features from tiny images.

\bibitem[{Kumar et~al.(2022)Kumar, Raghunathan, Jones, Ma, and Liang}]{kumarfine}
Kumar, A.; Raghunathan, A.; Jones, R.~M.; Ma, T.; and Liang, P. 2022.
\newblock Fine-Tuning can Distort Pretrained Features and Underperform Out-of-Distribution.
\newblock In \emph{International Conference on Learning Representations}.

\bibitem[{Li et~al.(2020)Li, Zhou, Xiong, and Hoi}]{li2020prototypical}
Li, J.; Zhou, P.; Xiong, C.; and Hoi, S.~C. 2020.
\newblock Prototypical contrastive learning of unsupervised representations.
\newblock \emph{arXiv preprint arXiv:2005.04966}.

\bibitem[{Li and Hoiem(2017)}]{li2017learning}
Li, Z.; and Hoiem, D. 2017.
\newblock Learning without forgetting.
\newblock \emph{IEEE transactions on pattern analysis and machine intelligence}, 40(12): 2935--2947.

\bibitem[{Lin et~al.(2023)Lin, Zhang, Feng, Li, and Ye}]{lin2023pcr}
Lin, H.; Zhang, B.; Feng, S.; Li, X.; and Ye, Y. 2023.
\newblock Pcr: Proxy-based contrastive replay for online class-incremental continual learning.
\newblock In \emph{Proceedings of the IEEE/CVF Conference on Computer Vision and Pattern Recognition}, 24246--24255.

\bibitem[{Liu et~al.(2020)Liu, Su, Liu, Schiele, and Sun}]{liu2020mnemonics}
Liu, Y.; Su, Y.; Liu, A.-A.; Schiele, B.; and Sun, Q. 2020.
\newblock Mnemonics training: Multi-class incremental learning without forgetting.
\newblock In \emph{Proceedings of the IEEE/CVF conference on Computer Vision and Pattern Recognition}, 12245--12254.

\bibitem[{Lu et~al.(2024)Lu, Feng, Yuan, Song, and Sun}]{lu2024revisiting}
Lu, A.; Feng, T.; Yuan, H.; Song, X.; and Sun, Y. 2024.
\newblock Revisiting Neural Networks for Continual Learning: An Architectural Perspective.
\newblock In \emph{IJCAI}, 4651--4659.

\bibitem[{Masana et~al.(2022)Masana, Liu, Twardowski, Menta, Bagdanov, and Van De~Weijer}]{masana2022class}
Masana, M.; Liu, X.; Twardowski, B.; Menta, M.; Bagdanov, A.~D.; and Van De~Weijer, J. 2022.
\newblock Class-incremental learning: survey and performance evaluation on image classification.
\newblock \emph{IEEE Transactions on Pattern Analysis and Machine Intelligence}, 45(5): 5513--5533.

\bibitem[{McCloskey and Cohen(1989)}]{mccloskey1989catastrophic}
McCloskey, M.; and Cohen, N.~J. 1989.
\newblock Catastrophic interference in connectionist networks: The sequential learning problem.
\newblock In \emph{Psychology of learning and motivation}, volume~24, 109--165. Elsevier.

\bibitem[{McDonnell et~al.(2024)McDonnell, Gong, Parvaneh, Abbasnejad, and van~den Hengel}]{mcdonnell2024ranpac}
McDonnell, M.~D.; Gong, D.; Parvaneh, A.; Abbasnejad, E.; and van~den Hengel, A. 2024.
\newblock Ranpac: Random projections and pre-trained models for continual learning.
\newblock \emph{Advances in Neural Information Processing Systems}, 36.

\bibitem[{Nagata and Hotta(2023)}]{nagata2023margin}
Nagata, K.; and Hotta, K. 2023.
\newblock Margin Contrastive Learning with Learnable-Vector for Continual Learning.
\newblock In \emph{Proceedings of the IEEE/CVF International Conference on Computer Vision}, 3570--3576.

\bibitem[{Oord, Li, and Vinyals(2018)}]{oord2018representation}
Oord, A. v.~d.; Li, Y.; and Vinyals, O. 2018.
\newblock Representation learning with contrastive predictive coding.
\newblock \emph{arXiv preprint arXiv:1807.03748}.

\bibitem[{Radford et~al.(2021)Radford, Kim, Hallacy, Ramesh, Goh, Agarwal, Sastry, Askell, Mishkin, Clark et~al.}]{radford2021learning}
Radford, A.; Kim, J.~W.; Hallacy, C.; Ramesh, A.; Goh, G.; Agarwal, S.; Sastry, G.; Askell, A.; Mishkin, P.; Clark, J.; et~al. 2021.
\newblock Learning transferable visual models from natural language supervision.
\newblock In \emph{International conference on machine learning}, 8748--8763. PMLR.

\bibitem[{Ridnik et~al.(2021)Ridnik, Ben-Baruch, Noy, and Zelnik-Manor}]{ridnik2021imagenet}
Ridnik, T.; Ben-Baruch, E.; Noy, A.; and Zelnik-Manor, L. 2021.
\newblock Imagenet-21k pretraining for the masses.
\newblock \emph{arXiv preprint arXiv:2104.10972}.

\bibitem[{Selvaraju et~al.(2017)Selvaraju, Cogswell, Das, Vedantam, Parikh, and Batra}]{selvaraju2017grad}
Selvaraju, R.~R.; Cogswell, M.; Das, A.; Vedantam, R.; Parikh, D.; and Batra, D. 2017.
\newblock Grad-cam: Visual explanations from deep networks via gradient-based localization.
\newblock In \emph{Proceedings of the IEEE international conference on computer vision}, 618--626.

\bibitem[{Smith et~al.(2023)Smith, Karlinsky, Gutta, Cascante-Bonilla, Kim, Arbelle, Panda, Feris, and Kira}]{smith2023coda}
Smith, J.~S.; Karlinsky, L.; Gutta, V.; Cascante-Bonilla, P.; Kim, D.; Arbelle, A.; Panda, R.; Feris, R.; and Kira, Z. 2023.
\newblock Coda-prompt: Continual decomposed attention-based prompting for rehearsal-free continual learning.
\newblock In \emph{Proceedings of the IEEE/CVF Conference on Computer Vision and Pattern Recognition}, 11909--11919.

\bibitem[{Snell, Swersky, and Zemel(2017)}]{snell2017prototypical}
Snell, J.; Swersky, K.; and Zemel, R. 2017.
\newblock Prototypical networks for few-shot learning.
\newblock \emph{Advances in neural information processing systems}, 30.

\bibitem[{Sun et~al.(2023)Sun, Zhou, Ye, and Zhan}]{sun2023pilot}
Sun, H.-L.; Zhou, D.-W.; Ye, H.-J.; and Zhan, D.-C. 2023.
\newblock PILOT: A Pre-Trained Model-Based Continual Learning Toolbox.
\newblock \emph{arXiv preprint arXiv:2309.07117}.

\bibitem[{Sun et~al.(2024)Sun, Zhou, Zhao, Gan, Zhan, and Ye}]{sun2024mos}
Sun, H.-L.; Zhou, D.-W.; Zhao, H.; Gan, L.; Zhan, D.-C.; and Ye, H.-J. 2024.
\newblock MOS: Model Surgery for Pre-Trained Model-Based Class-Incremental Learning.
\newblock \emph{arXiv preprint arXiv:2412.09441}.

\bibitem[{Tan et~al.(2024)Tan, Zhou, Xiang, Wang, Wu, and Li}]{tan2024semantically}
Tan, Y.; Zhou, Q.; Xiang, X.; Wang, K.; Wu, Y.; and Li, Y. 2024.
\newblock Semantically-Shifted Incremental Adapter-Tuning is A Continual ViTransformer.
\newblock In \emph{Proceedings of the IEEE/CVF Conference on Computer Vision and Pattern Recognition}, 23252--23262.

\bibitem[{Thengane et~al.(2022)Thengane, Khan, Hayat, and Khan}]{thengane2022continualclip}
Thengane, V.; Khan, S.; Hayat, M.; and Khan, F. 2022.
\newblock CLIP model is an Efficient Continual Learner.
\newblock \emph{arXiv:2210.03114}.

\bibitem[{Van~de Ven, Tuytelaars, and Tolias(2022)}]{van2022three}
Van~de Ven, G.~M.; Tuytelaars, T.; and Tolias, A.~S. 2022.
\newblock Three types of incremental learning.
\newblock \emph{Nature Machine Intelligence}, 4(12): 1185--1197.

\bibitem[{Van~der Maaten and Hinton(2008)}]{van2008visualizing}
Van~der Maaten, L.; and Hinton, G. 2008.
\newblock Visualizing data using t-SNE.
\newblock \emph{Journal of machine learning research}, 9(11).

\bibitem[{Wah et~al.(2011)Wah, Branson, Welinder, Perona, and Belongie}]{wah2011caltech}
Wah, C.; Branson, S.; Welinder, P.; Perona, P.; and Belongie, S. 2011.
\newblock The caltech-ucsd birds-200-2011 dataset.

\bibitem[{Wang et~al.(2023)Wang, Zhang, Su, and Zhu}]{wang2023comprehensive}
Wang, L.; Zhang, X.; Su, H.; and Zhu, J. 2023.
\newblock A comprehensive survey of continual learning: Theory, method and application.
\newblock \emph{arXiv preprint arXiv:2302.00487}.

\bibitem[{Wang et~al.(2022{\natexlab{a}})Wang, Zhang, Ebrahimi, Sun, Zhang, Lee, Ren, Su, Perot, Dy et~al.}]{wang2022dualprompt}
Wang, Z.; Zhang, Z.; Ebrahimi, S.; Sun, R.; Zhang, H.; Lee, C.-Y.; Ren, X.; Su, G.; Perot, V.; Dy, J.; et~al. 2022{\natexlab{a}}.
\newblock Dualprompt: Complementary prompting for rehearsal-free continual learning.
\newblock In \emph{European Conference on Computer Vision}, 631--648. Springer.

\bibitem[{Wang et~al.(2022{\natexlab{b}})Wang, Zhang, Lee, Zhang, Sun, Ren, Su, Perot, Dy, and Pfister}]{wang2022learning}
Wang, Z.; Zhang, Z.; Lee, C.-Y.; Zhang, H.; Sun, R.; Ren, X.; Su, G.; Perot, V.; Dy, J.; and Pfister, T. 2022{\natexlab{b}}.
\newblock Learning to prompt for continual learning.
\newblock In \emph{Proceedings of the IEEE/CVF conference on computer vision and pattern recognition}, 139--149.

\bibitem[{Wen et~al.(2024)Wen, Tan, Zheng, Xie, and Huang}]{wen2024provable}
Wen, Y.; Tan, Z.; Zheng, K.; Xie, C.; and Huang, W. 2024.
\newblock Provable Contrastive Continual Learning.
\newblock \emph{arXiv preprint arXiv:2405.18756}.

\bibitem[{Yan, Xie, and He(2021)}]{yan2021dynamically}
Yan, S.; Xie, J.; and He, X. 2021.
\newblock Der: Dynamically expandable representation for class incremental learning.
\newblock In \emph{Proceedings of the IEEE/CVF conference on computer vision and pattern recognition}, 3014--3023.

\bibitem[{Zenke, Poole, and Ganguli(2017)}]{zenke2017continual}
Zenke, F.; Poole, B.; and Ganguli, S. 2017.
\newblock Continual learning through synaptic intelligence.
\newblock In \emph{International conference on machine learning}, 3987--3995. PMLR.

\bibitem[{Zhai et~al.(2019)Zhai, Puigcerver, Kolesnikov, Ruyssen, Riquelme, Lucic, Djolonga, Pinto, Neumann, Dosovitskiy et~al.}]{zhai2019large}
Zhai, X.; Puigcerver, J.; Kolesnikov, A.; Ruyssen, P.; Riquelme, C.; Lucic, M.; Djolonga, J.; Pinto, A.~S.; Neumann, M.; Dosovitskiy, A.; et~al. 2019.
\newblock A large-scale study of representation learning with the visual task adaptation benchmark.
\newblock \emph{arXiv preprint arXiv:1910.04867}.

\bibitem[{Zhang, Song, and Tao(2022)}]{zhang2022hierarchical}
Zhang, X.; Song, D.; and Tao, D. 2022.
\newblock Hierarchical prototype networks for continual graph representation learning.
\newblock \emph{IEEE Transactions on Pattern Analysis and Machine Intelligence}, 45(4): 4622--4636.

\bibitem[{Zhang et~al.(2022)Zhang, Yin, Shao, and Liu}]{zhang2022benchmarking}
Zhang, Y.; Yin, Z.; Shao, J.; and Liu, Z. 2022.
\newblock Benchmarking omni-vision representation through the lens of visual realms.
\newblock In \emph{European Conference on Computer Vision}, 594--611. Springer.

\bibitem[{Zhao et~al.(2021)Zhao, Wang, Fu, Wu, and Li}]{zhao2021memory}
Zhao, H.; Wang, H.; Fu, Y.; Wu, F.; and Li, X. 2021.
\newblock Memory-efficient class-incremental learning for image classification.
\newblock \emph{IEEE Transactions on Neural Networks and Learning Systems}, 33(10): 5966--5977.

\bibitem[{Zhou et~al.(2024{\natexlab{a}})Zhou, Cai, Ye, Zhan, and Liu}]{zhou2024revisiting}
Zhou, D.-W.; Cai, Z.-W.; Ye, H.-J.; Zhan, D.-C.; and Liu, Z. 2024{\natexlab{a}}.
\newblock Revisiting class-incremental learning with pre-trained models: Generalizability and adaptivity are all you need.
\newblock \emph{International Journal of Computer Vision}, 1--21.

\bibitem[{Zhou et~al.(2024{\natexlab{b}})Zhou, Sun, Ning, Ye, and Zhan}]{zhou2024continual}
Zhou, D.-W.; Sun, H.-L.; Ning, J.; Ye, H.-J.; and Zhan, D.-C. 2024{\natexlab{b}}.
\newblock Continual learning with pre-trained models: A survey.
\newblock \emph{arXiv preprint arXiv:2401.16386}.

\bibitem[{Zhou et~al.(2023)Zhou, Wang, Ye, and Zhan}]{zhou2023model}
Zhou, D.-W.; Wang, Q.-W.; Ye, H.-J.; and Zhan, D.-C. 2023.
\newblock A Model or 603 Exemplars: Towards Memory-Efficient Class-Incremental Learning.
\newblock In \emph{ICLR}.

\bibitem[{Zhou et~al.(2022)Zhou, Liu, Qiao, Xiang, and Loy}]{zhou2022domain}
Zhou, K.; Liu, Z.; Qiao, Y.; Xiang, T.; and Loy, C.~C. 2022.
\newblock Domain generalization: A survey.
\newblock \emph{IEEE transactions on pattern analysis and machine intelligence}, 45(4): 4396--4415.

\end{thebibliography}

\end{document}